\pdfoutput=1

\documentclass[11pt]{article}

\usepackage{latex/acl}
\usepackage{times}  
\usepackage{helvet}  
\usepackage{courier}  
\usepackage{graphicx} 
\usepackage{natbib}  
\usepackage{caption} 

\usepackage{times}
\usepackage{latexsym}

\usepackage{lscape}
\usepackage{amsmath}
\usepackage{algorithm}
\usepackage{algorithmic}
\usepackage{booktabs}
\usepackage{subcaption}
\usepackage{amssymb}
\usepackage{amsmath}
\usepackage{enumitem}
\usepackage{array}
\usepackage{soul}
\usepackage{xcolor}
\usepackage{array}
\usepackage{multirow}
\usepackage{multicol}
\usepackage{soul}

\DeclareMathOperator*{\minimize}{minimize}

\ExplSyntaxOn
\NewDocumentCommand{\longdash}{ O{2} }
 {
  --\prg_replicate:nn { #1 - 1 } { \negthinspace -- }
 }
\ExplSyntaxOff

\usepackage[T1]{fontenc}

\usepackage[utf8]{inputenc}

\usepackage{microtype}

\usepackage{inconsolata}
\usepackage{array}

\usepackage{wasysym}

\title{A Curious Case of Searching for the Correlation between Training Data and Adversarial Robustness of Transformer Textual Models}

\author{
    Cuong Dang\textsuperscript{\rm 1},
    Dung D. Le\textsuperscript{\rm 2},
    Thai Le\textsuperscript{\rm 3},\\
    \textsuperscript{\rm 1}FPT Software AI Center, Vietnam\\
    \textsuperscript{\rm 2}College of Engineering and Computer Science, VinUniversity, Vietnam\\
    \textsuperscript{\rm 3}Department of Computer Science, Indiana University, USA\\
    cuongdc10@fpt.com, 
    dung.ld@vinuni.edu.vn,
    tle@iu.edu,
}

\begin{document}
\maketitle
\begin{abstract}
\vspace{-10pt}
Existing works have shown that fine-tuned textual transformer models achieve state-of-the-art prediction performances but are also vulnerable to adversarial text perturbations. Traditional adversarial evaluation is often done \textit{only after} fine-tuning the models and ignoring the training data. In this paper, we want to prove that there is also a strong correlation between training data and model robustness. To this end, we extract 13 different features representing a wide range of input fine-tuning corpora properties and use them to predict the adversarial robustness of the fine-tuned models. Focusing mostly on encoder-only transformer models BERT and RoBERTa with additional results for BART, ELECTRA, and GPT2, we provide diverse evidence to support our argument. First, empirical analyses show that (a) extracted features can be used with a lightweight classifier such as Random Forest to predict the attack success rate effectively, and (b) features with the most influence on the model robustness have a clear correlation with the robustness. Second, our framework can be used as a fast and effective additional tool for robustness evaluation since it (a) saves 30x-193x runtime compared to the traditional technique, (b) is transferable across models, (c) can be used under adversarial training, and (d) robust to statistical randomness. Our code is publicly available at \url{https://github.com/CaptainCuong/RobustText_ACL2024}.
\end{abstract}



\vspace{-10pt}
\section{Introduction}
\vspace{-5pt}
Pre-trained transformer models such as BERT~\cite{devlin-etal-2019-bert} and RoBERTa~\cite{liu2019RoBERTa} have recently demonstrated superior performance in various downstream NLP classification tasks. However, they are also vulnerable to adversarial text attacks~\cite{sun2020adv,Jin_Jin_Zhou_Szolovits_2020}, which aim to generate adversarial examples by applying imperceptible perturbations to input texts such that the resulting examples cause a target text classifier to make incorrect predictions~\cite{goodfellow2015explaining}. This makes the robustness of transformer models against adversarial attacks crucial, especially in high-stake domains such as banking, law, and content moderation~\cite{rodriguez2021matter,sanz2022disconnect,ashley2019brief} where susceptibility to such attacks can result in detrimental consequences such as giving out high-risk loans, wrongful indictments and enabling hate speech and disinformation. Thus, ML practitioners must ensure their models are robust against text perturbations before deploying them to the public.
\begin{figure}[tb]
    \vspace{-5pt}
    \centering
    \includegraphics[width=\linewidth]{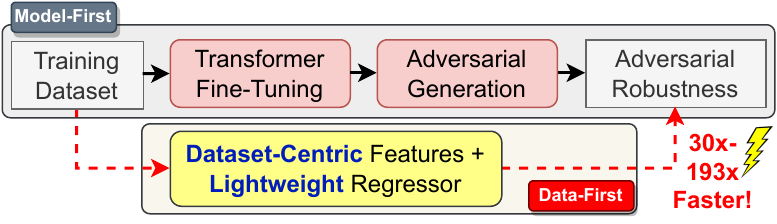}
    \caption{A novel attempt to bypass both model fine-tuning and adversarial generation and correlate adversarial robustness \textit{directly from the training dataset}, potentially saving 30x–193x of runtime.}
    \label{intro_diagram}
    \vspace{-20pt}
\end{figure}
To achieve this, existing works have proposed several ways to benchmark and analyze the robustness of perturbations of transformer models. In general, they often take a \textbf{model-first approach}--i.e., assuming that the model itself, such as its architecture or loss function formulation, is mainly responsible for its adversarial vulnerability and aiming to understand what kinds of changes in a model would shift its adversarial robustness~\cite{mao2022towards,zhang2022interpreting,zhang2022improving,han2024designing}. Particularly, this approach iteratively makes a controlled alternation in the model--e.g., changing the architecture type, experimenting with novel attention layers, adding noises to the embeddings, etc., and then fine-tune the new model on \textit{the same fine-tuning dataset}, followed by generating adversarial examples and benchmarking the model using the generated examples.
\begin{figure*}[tb]
    \centering
    \includegraphics[width=0.9\textwidth]{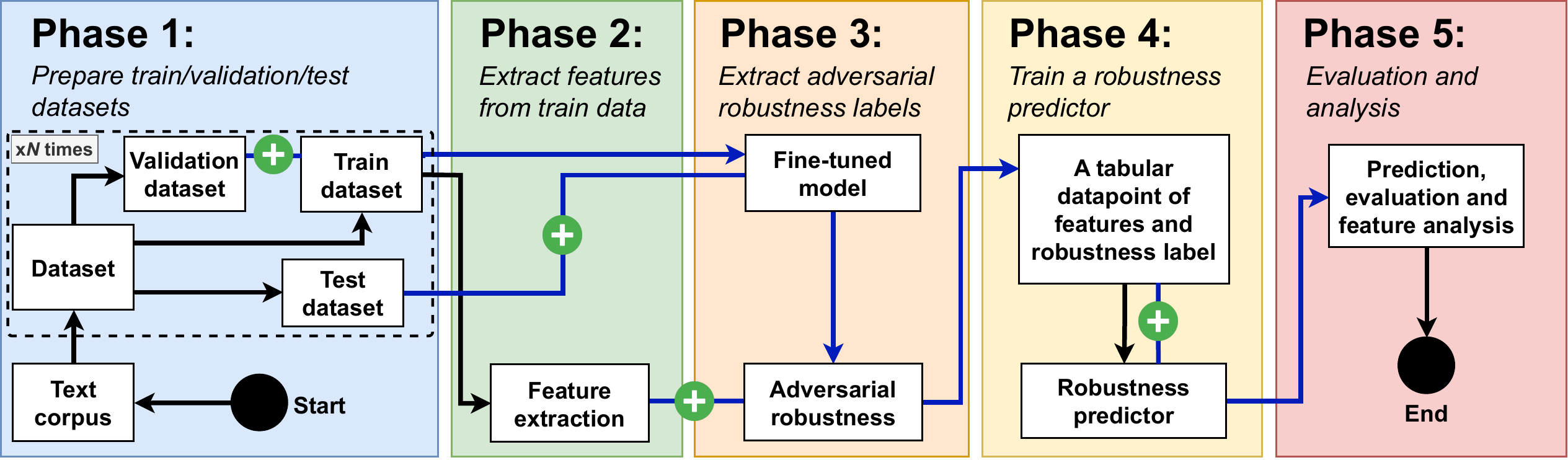}
    \caption{An illustrative overview of our framework for data-first adversarial robustness analysis. Black and blue arrows take one and two previous input(s), respectively, and return an output.
    }
    \label{flow_chart}
    \vspace{-0.2in}
\end{figure*}
Although this model-first approach has resulted in several useful insights in practice, it assumes that adversarial robustness can only be evaluated \textit{only after} a model has already been fine-tuned and adversarial examples have been generated. This approach tends to isolate the effects of the fine-tuning dataset and hence does not provide many insights on how such training data affect a model's robustness, such as ``how the distribution of fine-tuning texts' embeddings and labels affect a model's robustness?'', ``how do the unique vocabulary and lengths of fine-tuning texts correlate with a model's robustness?'', etc.  Exploring the relationship between fine-tuning data and model robustness promises to open up data-centric research directions to improve model robustness, introduced in detail in Sec. \ref{future}. Therefore, in this paper, we propose a \textit{global interpretation framework}, as shown in Fig. \ref{intro_diagram}, to investigate whether there is a strong direct correlation between fine-tuning data and model robustness and interpret the features of fine-tuning data that have the greatest influences.

To this end, we take a different approach from model-first analysis and propose to analyze the adversarial robustness from a \textbf{data-first approach}. To do this, we extract 13 different features that comprehensively capture several important properties of not individual training examples but of the fine-tuning dataset as a whole. Then, via regression analysis, we \textit{attempt} to correlate them with the adversarial robustness of the models \textbf{to be fine-tuned} on the dataset measured as the average attack success rates (ASRs) of 4 representative text perturbation methods on an unseen test set.


To demonstrate one application of such novel analysis, we also try utilizing our interpretation framework to \textbf{estimate the adversarial robustness} of transformer classifiers even \ul{before} they are fine-tuned and \ul{without} the need to generate adversarial examples, only by analyzing their fine-tuning dataset. This approach is 30-193 times faster than the traditional method, as shown in Fig. \ref{intro_diagram}.

{\bf Contributions} of our paper are as follows.
\begin{itemize}[leftmargin=\dimexpr\parindent-0.2\labelwidth\relax,noitemsep]
    \vspace{-10pt}
    \item To the best of our knowledge, this is the first paper to analyze and investigate a comprehensive correlation between fine-tuning data and model robustness with a taxonomy of 13 dataset-level indicators,
    \item \textit{As an application}, we demonstrate that this novel analysis also enables a Random Forest predictor to effectively evaluate the adversarial robustness of BERT and RoBERTA with the averaged mean absolute errors (MAEs) ranging in 0.025--0.176 for both in-domain and out-of-domain prediction, 
    \item Our framework can also be used as a fast tool to evaluate the robustness of transformer-based text classifiers, which (i) is 30x-193x faster than the usual procedure, (ii) can be used under an adversarial training setting, (iii) transferable between transformer-based models, and (iv) robust to statistical randomness.
\end{itemize}

\section{Problem Formulation}
\vspace{-5pt}
We propose to develop a function $\mathcal{G}^f_\theta(\mathcal{D})$ parameterized by $\theta$ that can effectively approximate the adversarial robustness of a pre-trained transformer-based classification model $f$ when it is fine-tuned by \textit{an input} training dataset $\mathcal{D}$. In other words, $\mathcal{G}^f_\theta(\mathcal{D})$ estimates the difference between predictions on examples of a clean, unseen test set $\mathcal{D}^*$ ($\mathcal{D}\cap\mathcal{D}^*{=}\emptyset$) that is \textit{drawn from the same distribution with $\mathcal{D}$ and is sufficiently large} and on their corresponding adversarial examples. Let's denote $\mathbf{R}(f, \mathcal{D}, \mathcal{D}^*)$ such adversarial robustness, we have: 
\setlength{\belowdisplayskip}{3pt}
\setlength{\abovedisplayskip}{3pt}
\begin{equation}
    \mathbf{R}(f, \mathcal{D}, \mathcal{D}^*) = \frac{1}{|\mathcal{D}^*|}{\sum_{x\in\mathcal{D}^*}} d(f_{\mathcal{D}}(x), f_{\mathcal{D}}(x+\delta)),
\end{equation}
\noindent where $\delta$ is an adversarial perturbation and $d(\cdot)$ is a metric such as \textit{attack success rate} as often adopted in existing literature. Since $\mathcal{D}^*$ is sufficiently large, we assume to observe only a small variance among the adversarial robustness measured on different randomly sampled $\mathcal{D}^*$, drawn from the same distribution as $\mathcal{D}$. Hence, we simplify the adversarial robustness to be estimated as $\mathbf{R}(f, \mathcal{D}){\approx}\mathbf{R}(f, \mathcal{D}, \mathcal{D}^*)$ with any arbitrary $\mathcal{D}^*$.

To train $\mathcal{G}^f_\theta$, which is specific to the model type $f$ such as BERT or RoBERTa, we can then formulate this as a regression prediction problem and minimize the $L_2$ loss for an arbitrary fine-tuning dataset $\mathcal{D}$ as follows.
\begin{equation}
    \minimize_{\theta} \mathcal{L}_\mathcal{D} = ||\mathcal{G}_{\theta}^f({\mathcal{D}}) - \mathbf{R}(f, \mathcal{D})||^2_2,
\end{equation}
\noindent where $\mathcal{L}_{\mathcal{D}}$ is then the loss for one fine-tuning corpora $\mathcal{D}$. To effectively train $\mathcal{G}$ that can approximate adversarial robustness for \textit{any unseen fine-tuning corpus}, we will need to optimize such loss function not for one but $N$ training corpus $\pmb{\mathcal{Q}}{=}\{\mathcal{Q}_1, \mathcal{Q}_2,..\mathcal{Q}_N\}$, resulting in the final objective with \textit{mean square error (MSE)} loss function:
\begin{equation}
    \mathcal{L} = \frac{1}{N}\sum_{\mathcal{Q}\in\pmb{\mathcal{Q}}}\mathcal{L}_{\mathcal{Q}}
\end{equation}
In this work, we want to evaluate $\mathcal{G}_{\theta}^f$ in two prediction scenarios, namely \textit{interpolation} and \textit{extrapolation}. In \textit{(1) interpolation or in-distribution evaluation}, we want to validate $\mathcal{G}_{\theta}^f$ on a fine-tuning corpus that is similar to one of the corpus included in $\pmb{\mathcal{Q}}$ that the model $\mathcal{G}_{\theta}^f$ has been trained on. This is also the standard evaluation setting in a typical machine learning problem. In \textit{(2) extrapolation or out-of-distribution evaluation}, we want to validate $\mathcal{G}_{\theta}^f$ on a dataset that is very different from corpus included in $\pmb{\mathcal{Q}}$--e.g., training on sentiment classification datasets and evaluating on a non-sentiment dataset such as Q\&A or fakenews detection.
\begin{figure}[tb]
\centering
\includegraphics[width=\linewidth]{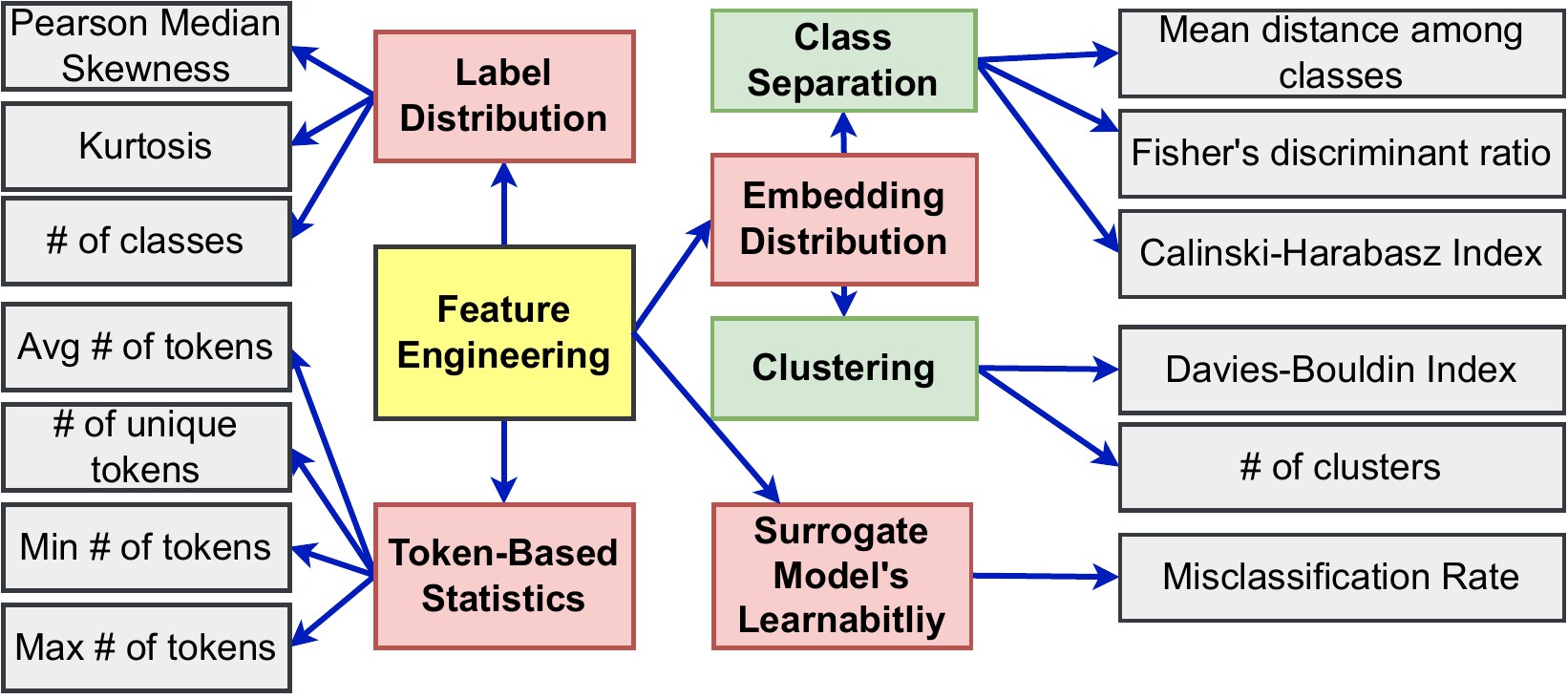}
\caption{Taxonomy of 13 predictive features (gray) categorized into groups (red) and sub-groups (green).
}
\label{taxanomy}
\vspace{-10pt}
\end{figure}

\vspace{-5pt}
\section{Method}
\vspace{-5pt}
\noindent \textbf{Overview.} 
Our goal is to create a regression dataset that includes (1) the features of the several smaller datasets and (2) their adversarial robustness--i.e., \textit{attack success rate (ASR)} of a transformer-based model $f$ fine-tuned on each of them. Then, we can use regression ML algorithms to predict such adversarial robustness and then analyze the influence of those features on the robustness of the model. Fig. \ref{flow_chart} illustrates the entire framework of five sequential phases.

\begin{algorithm}[tb]
\caption{\textit{Data Preparation Pseudo-code}}
\footnotesize
\label{alg:algorithm}
\textbf{Input}: Set of text corpus $\pmb{\mathcal{D}}$, training sample size $N$\\
\textbf{Output}: Final datasets $\mathcal{\pmb{Q}}$ to be used for training/validation/testing\\
\textbf{Initialize:} $\mathcal{\pmb{Q}} \leftarrow \emptyset$, $i\leftarrow 0$
\begin{algorithmic}[1]
\FOR{corpus $d$ in $\pmb{\mathcal{D}}$}
\STATE Randomly sample $S^{d}_\mathrm{test}{\in}S$ 
\ENDFOR
\FOR{$i$ in [1..N]}
\STATE Randomly a sample corpus $d$ from $\pmb{\mathcal{D}}$.
\STATE Randomly sample $S^i_\mathrm{train}$, $S^i_\mathrm{val}$ from $d$ such that
\STATE $\quad S^i_\mathrm{train}{\cap}S^{d}_\mathrm{test}{=}\emptyset$; $S^i_\mathrm{val}{\cap}S^{d}_\mathrm{test}{=}\emptyset$; $S^i_\mathrm{train}{\cap}S^i_\mathrm{val}{=}\emptyset$
\STATE $\mathcal{\pmb{Q}} \leftarrow \mathcal{\pmb{Q}} \cup (S^i_\mathrm{train}, S^i_\mathrm{val}, S^{d}_\mathrm{test})$  
\ENDFOR
\STATE \textbf{return} $\mathcal{\pmb{Q}}$
\end{algorithmic}
\end{algorithm}

\vspace{-5pt}
\subsection{Phase 1: Data Preparation.}
\label{prepare_subdataset}
\vspace{-5pt}
Our dataset preparation pipeline starts with set $\mathcal{\pmb{D}}$, which includes 9 diverse and publicly available NLP classification corpus. Different from a typical ML problem, in this work, each training example is a dataset and not a single text. Hence, we proposed a data splitting strategy as shown in Algorithm \ref{alg:algorithm}. For each text corpus $d \in \mathcal{\pmb{D}}$, we first randomly sample a test set of size $K$ to be used for calculating the attack success rate--i.e., adversarial robustness, as prediction labels in Phase 3 (Fig. \ref{flow_chart}) (Alg. \ref{alg:algorithm}, Ln. 1--3). To sample one instance in our final dataset, we first randomly pick a text corpus $d \in \mathcal{\pmb{D}}$ and randomly sample from it a small train and validation set of size $9*K$ and $K$ to achieve a 9:1 ratio between train and validation set, then pair them with the fixed test set previously sampled for $d$ (Alg. \ref{alg:algorithm}, Ln. 5--7). We repeat such process $N$ (Alg. \ref{alg:algorithm} Ln. 4--8) times to sample $N$ total triplets of \emph{non-overlapping} train, validation, and test sets. 

\vspace{-5pt}
\subsection{Phase 2: Feature Engineering.}
\label{extract_features}
\vspace{-5pt}
This phase extracts a total of 13 features that capture different aspects of each fine-tuning dataset (Fig. \ref{flow_chart}) for robustness prediction afterward. The features are categorized into 4 aspects, namely \textit{Embedding Distribution}, \textit{Label Distribution}, \textit{Weak Model's Learnability}, and \textit{Dataset Statistics}. Within each aspect, we develop several quantitative predictive indicators as summarized in Fig. \ref{taxanomy}. Our goal is to investigate the influence of these features on the adversarial robustness of the fine-tuned transformer-based models. 

\begin{itemize}[leftmargin=\dimexpr\parindent-0.5\labelwidth\relax,noitemsep]
    \vspace{-5pt}
    \item \textbf{Embedding Distribution.} Inspired by \cite{yu2018interpreting} which shows the influence of input space on the adversarial robustness of transformer-based models, we propose to use several indicators that summarize how closely the included texts are distributed in the embedding space. They are \textit{(1) mean distance among classes (MD)}, \textit{(2) Fisher's discriminant ratio (F)}, \textit{(3) Calinski-Harabasz Index (CHI)}, \textit{(4) Davies-Bouldin Index (DBI)} and \textit{(4) number of clusters (\# of clusters)}. To do this, we use the Universal Sentence Encoding ~\cite{cer2018universal} to encode the sentences in each fine-tuning dataset into embedding vectors. 
    
    \item \textbf{Label Distribution.} Fine-tuning datasets with a skewed or peaked label distribution can lead to biased predictions, especially for complex transformer-based models that are prone to overfitting to the majority class and lead to poor generalization. 
    Thus, we adopt several indicators to quantify the skewness and peakedness of input labels, including \textit{(1) Pearson Median Skewness (PMS)}, \textit{(2) Kurtosis (Kurt)}. 
    Furthermore, we include the number of labels as a feature so that robustness prediction may be tailored to specific tasks.
    \item \textbf{Surrogate Model's Learnability.} Inspired by \cite{zhang2022interpreting}, we assume that the predictive performance of a weak model on the fine-tuning dataset can also inform about potential predictive biases that will also transfer to transformer-based models. We coin this feature \textit{Misclassification Rate (MCR)}. Intuitively, a surrogate model with good predictive performance makes it more likely that a fine-tuned transformer model will also achieve similar or even better generalization. Conversely, a surrogate model with poor predictive performance provides a quick sanity check for potential biases in the fine-tuning dataset--e.g., inconsistent, noisy, or skewness in labels, which will eventually lead to poor generalization of the fine-tuned transformer-based model. Particularly, we use a character-based CNN classifier that is smaller than a typical transformer-based model as the surrogate model. Such a model is more computationally efficient during training and inference, and more powerful than traditional ML classifiers such as Naive Bayes or Decision Tree.

    \item \textbf{Token-Based Statistics.} The length of input text and typos affect the robustness of the transformer-based model~\cite{jia-liang-2017-adversarial,sun2020adv}. Hence, we examine the influence of some summary statistics of the dataset on the robustness of the transformer-based model, namely the \textit{(1) average number of tokens (avg. \# tokens)}, \textit{(2) the minimum number of tokens (min \# tokens)}, \textit{(3) the maximum number of tokens (max \# tokens)}, and \textit{(4) the number of unique tokens (\# unique tokens)}. These statistics reflect the types of texts where a fine-tuned transformer-based model has observed and thereby informs its performance when dealing with unseen, adversarial examples. 
    \vspace{-12pt}
\end{itemize}
\begin{figure}[tb]
    \centering
    \includegraphics[width=\linewidth]{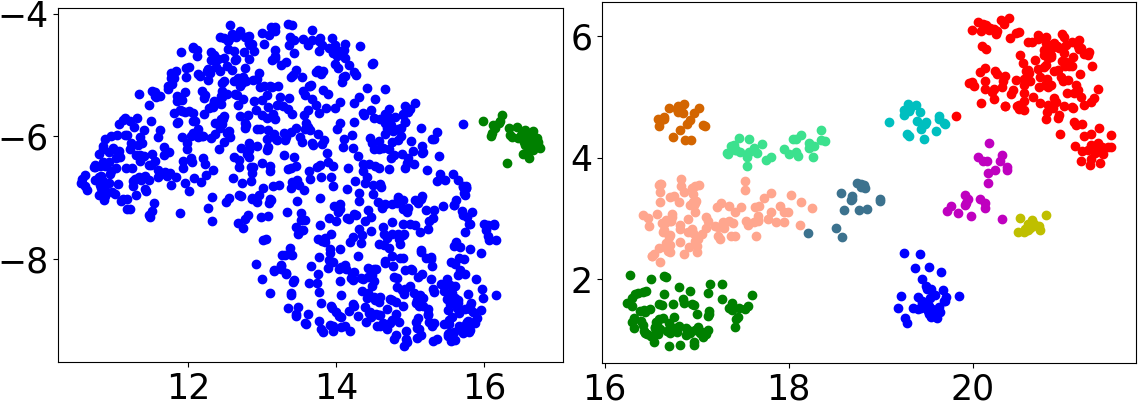}
    \caption{Embeddings of two fine-tuning datasets projected on a 2D space by t-SNE \cite{van2008visualizing}. Dataset with more separated clusters (right) results in a fine-tuned model that is \textit{more vulnerable} to adversarial perturbations.
    }
    \label{embed}
    \vspace{-15pt}
\end{figure}
\vspace{-5pt}
\subsection{Phase 3: Extract Adversarial Robustness as Regression Labels.}\label{pred_asr}
\vspace{-5pt}
Phase 3 aims to predict the adversarial robustness of the model after fine-tuning the datasets prepared in Phase 1 (Fig. \ref{flow_chart}). After extracting the features of the fine-tuning data $\mathcal{S}_{train}$, we fine-tune a transformer-based classifier $f(\cdot)$ on the training dataset $\mathcal{S}_{train}$ and validate on $\mathcal{S}_{val}$. Then, the adversarial robustness of $f(\cdot)$ will be extracted by averaging the attack success rates of four text perturbation methods used to attack $f(\cdot)$. They include one character-level attacker DeepWordBug~\cite{gao2018black} and three word-level attackers BERT-Attack~\cite{li-etal-2020-bert-attack}, PWWS~\cite{ren-etal-2019-generating}, and TextFooler~\cite{Jin_Jin_Zhou_Szolovits_2020}. These four attackers are both standard benchmark text perturbation methods in the literature and represent diverse attack methods in practice. 

\vspace{-5pt}
\subsection{Phase 4: Regression Analysis through Adversarial Robustness Estimation.}
\vspace{-5pt}
Phase 4 aims to train a regression classifier $\mathcal{G}^f_\theta(\cdot)$ that inputs the engineered features of a \textit{fine-tuning dataset} and predicts the adversarial robustness, measured by ASR, of a corresponding fine-tuned transformer-based architecture $f$, for $f$ is either BERT or RoBERTa (Fig. \ref{flow_chart}). Phase 1, 2, and 3 have provided us with a \textit{tabular training dataset} total of $N$ data points, each of which contains the engineered features of each small fine-tuning dataset $\mathcal{S}_\mathrm{train}$ and its corresponding ASR on unseen $\mathcal{S}_\mathrm{test}$ of a fine-tuned transformer-based model. We adopt three popular ML models for predictor $\mathcal{G}^f_\theta$, namely \textit{Gradient Boosting}, \textit{Linear Regression}, and \textit{Random Forest}. These predictors are computationally efficient and achieve competitive predictive performance compared to advanced deep models on tabular datasets~\cite{grinsztajn2022tree}.
\vspace{-5pt}
\subsection{Phase 5: Evaluation and Analysis.}
\vspace{-5pt}
To evaluate and gain meaningful insights into the trained predictor $\mathcal{G}^f_\theta$, we report results and carry out analyses as follows.
\begin{itemize}[leftmargin=\dimexpr\parindent-0.7\labelwidth\relax,noitemsep]
    \vspace{-10pt}
    \item \textbf{Runtime:} We compare the runtime of our framework over the conventional adversarial robustness measurement approach which requires both fine-tuning a model and generating adversarial examples.
    \item \textbf{Prediction Performance:} We evaluate our framework under two inference scenarios, namely \textit{interpolation} and \textit{extrapolation}. Interpolation is the process of estimating ASRs within the domain of observed data points while extrapolation, conversely, is a prediction of ASRs on out-of-domain data. Although extrapolation evaluation is more challenging, it is more practical as we want to evaluate how well our regression predictor $\mathcal{G}^f_\theta$ performs on a corpus that it does not see during training.
    \item \textbf{Feature Analysis:} We adopt the \textit{Permutation Feature Importance} and \textit{Accumulated Local Effects} technique to estimate and analyze the influence of engineered features on $\mathcal{G}^f_\theta$'s ASR predictions--i.e., how their values correlate with the predicted adversarial robustness, and their importance rankings.
    \item \textbf{Prediction under Adversarial Training:} We evaluate our adversarial robustness predictor under adversarial training setting~\cite{goodfellow2015explaining}. Adversarial training is a popular technique that helps improve a model's robustness by training a model with additional adversarial perturbations. This means that a good predictor $\mathcal{G}^f_\theta$ is expected to consistently output smaller ASRs, and hence informing a more robust model, under this setting.
    \item \textbf{Prediction Transferability:} Transformer-based models are well-known for robustness transferability. To put it another way, their robustness is quite the same. Thus, we expect our robustness predictor to work on other untrained transformer models at an acceptable level.
    \item \textbf{Prediction Consistency:} Since ASR is a statistical metric, randomness is inevitable. We examine whether these statistical labels affect the performance of our robustness prediction.
\end{itemize}

\vspace{-10pt}
\section{Related Work}
\vspace{-5pt}
\noindent \textbf{Adversarial Attacks in NLP.} 
The general framework for adversarial attacks on a sentence includes two steps: (1) choosing which words in the sentence a target text classifier is most vulnerable to and (2) replacing them with a candidate such that the prediction label crosses the original prediction. Thus, most of the attack methods differ in how they come up with new replacements, with the majority of them using word-level perturbation strategies such as via word-substitution (~\cite{li-etal-2020-bert-attack,Jin_Jin_Zhou_Szolovits_2020,ren-etal-2019-generating}) or character-level attack such as via swapping and deleting characters within an original word (~\cite{gao2018black,li2019textbugger}). While one can choose a set of random words in a sentence to perturb, existing works also propose several optimization schemes such as \textit{greedy search} or \textit{genetic algorithm} to select the optimal words to perturb and also their replacements. Although these mechanisms help maximize the changes in the target classifier's behaviors while still preserving the sentence's original semantic meaning, the fact that they work with discrete NLP domain induces a substantial additional computational cost due to the need to continuously ping the target model for fine-tuning their perturbations, often on one token at a time.
\noindent \textbf{Interpreting the Adversarial Robustness of Models.}
\cite{zhang2022improving} claimed that a lack of model robustness is caused by non-robust features. As a result, they improved text classification models by including a bottleneck layer in their architectures to eliminate the effects of low-quality features. Moreover, \cite{han2024designing} attributed the non-robust transformer models to outliers, and presented a resilient framework called transformer-RKDE by replacing the dot-product attention with attention deriving from robust kernel density estimators. In addition to these works that focus more on model architectures, works such as \cite{jia-liang-2017-adversarial} focused more on drawing the relationship between specific linguistic patterns and the adversarial robustness, but only on unseen test sentences during inference. Distinguished from these works, we emphasize and analyze the role of the fine-tuning dataset during model training on adversarial robustness and isolate the effects of the model architecture and inference inputs.
\newcolumntype{H}{>{\setbox0=\hbox\bgroup}c<{\egroup}@{}}
\renewcommand{\tabcolsep}{3pt}
\begin{table}[tb]
\centering
\footnotesize
\begin{tabular}{llHHcHHc}
\toprule
{} & \multicolumn{1}{c}{\multirow{1}{*}{\textbf{METRIC}}} & \multicolumn{3}{c}{\textbf{INTERPOLATION}} & \multicolumn{3}{c}{\textbf{EXTRAPOLATION}} \\
\cmidrule(lr){3-5} \cmidrule(lr){6-8}
\cmidrule(lr){1-8}
\multirow{5}{*}{\rotatebox[origin=c]{90}{\textbf{BERT}}}        & RMSE${\downarrow}$ & $0.059\pm0.000$ & $0.072\pm0.000$ & $0.055\pm0.000$ & $0.169\pm0.005$ & $0.063\pm0.003$ & $0.063\pm0.001$ \\
\textbf{}        & $R^2$${\uparrow}$   & $0.892\pm0.006$ & $0.841\pm0.007$ & $0.904\pm0.005$ & $0.394\pm0.177$ & $0.871\pm0.122$ & $0.885\pm0.033$ \\
{}    & MAE${\downarrow}$  & $0.040\pm0.000$ & $0.053\pm0.000$ & $0.037\pm0.000$ & $0.128\pm0.006$ & $0.040\pm0.001$ & $0.045\pm0.000$ \\
\textbf{}        & EVS${\uparrow}$  & $0.895\pm0.006$ & $0.846\pm0.007$ & $0.907\pm0.005$ & $0.522\pm0.089$ & $0.892\pm0.060$ & $0.908\pm0.021$ \\
\textbf{}        & MAPE${\downarrow}$ & $0.077\pm0.001$ & $0.101\pm0.001$ & $0.071\pm0.000$ & $0.278\pm0.020$ & $0.086\pm0.006$ & $0.102\pm0.004$ \\ 
\cmidrule(lr){1-8}
\multirow{5}{*}{\rotatebox[origin=c]{90}{\textbf{RoBERTa}}}        & RMSE${\downarrow}$ & $0.037\pm0.000$ & $0.056\pm0.000$ & $0.031\pm0.000$ & $0.206\pm0.005$ & $0.073\pm0.003$ & $0.061\pm0.001$ \\
\textbf{}        & $R^2$${\uparrow}$   & $0.959\pm0.000$ & $0.907\pm0.001$ & $0.972\pm0.000$ & $0.139\pm0.145$ & $0.829\pm0.205$ & $0.900\pm0.019$ \\
\textbf{}    & MAE${\downarrow}$  & $0.028\pm0.000$ & $0.044\pm0.000$ & $0.025\pm0.000$ & $0.176\pm0.006$ & $0.042\pm0.001$ & $0.044\pm0.000$ \\
\textbf{}        & EVS${\uparrow}$ & $0.961\pm0.000$ & $0.911\pm0.001$ & $0.972\pm0.000$ & $0.309\pm0.109$ & $0.846\pm0.153$ & $0.922\pm0.010$ \\
\textbf{}        & MAPE${\downarrow}$ & $0.054\pm0.000$ & $0.083\pm0.000$ & $0.048\pm0.000$ & $0.385\pm0.032$ & $0.083\pm0.003$ & $0.095\pm0.004$ \\ 
\cmidrule(lr){1-8}
\multirow{5}{*}{\rotatebox[origin=c]{90}{\textbf{ELECTRA}}}        & RMSE${\downarrow}$ & $0.107\pm0.001$ & $0.084\pm0.002$ & $0.070\pm0.001$ & $0.135\pm0.004$ & $0.148\pm0.009$ & $0.073\pm0.000$ \\
\textbf{}        & $R^2$${\uparrow}$   & $0.411\pm0.492$ & $0.635\pm0.194$ & $0.686\pm0.490$ & $0.450\pm0.240$ & $0.348\pm0.694$ & $0.864\pm0.007$ \\
{}    & MAE${\downarrow}$  & $0.083\pm0.001$ & $0.057\pm0.001$ & $0.047\pm0.000$ & $0.100\pm0.005$ & $0.064\pm0.000$ & $0.039\pm0.000$ \\
\textbf{}        & EVS${\uparrow}$  & $0.505\pm0.293$ & $0.677\pm0.152$ & $0.729\pm0.326$ & $0.513\pm0.174$ & $0.361\pm0.671$ & $0.870\pm0.005$ \\
\textbf{}        & MAPE${\downarrow}$ & $0.151\pm0.006$ & $0.105\pm0.004$ & $0.084\pm0.003$ & $0.180\pm0.012$ & $0.129\pm0.002$ & $0.077\pm0.000$ \\ 
\cmidrule(lr){1-8}
\multirow{5}{*}{\rotatebox[origin=c]{90}{\textbf{GPT2}}}        & RMSE${\downarrow}$ & $0.093\pm0.002$ & $0.026\pm0.000$ & $0.025\pm0.000$ & $0.110\pm0.002$ & $0.147\pm0.009$ & $0.078\pm0.000$ \\
\textbf{}        & $R^2$${\uparrow}$   & $-0.468\pm37.303$ & $0.888\pm0.105$ & $0.890\pm0.106$ & $0.523\pm0.135$ & $-0.013\pm3.437$ & $0.794\pm0.005$ \\
\textbf{}    & MAE${\downarrow}$  & $0.067\pm0.001$ & $0.020\pm0.000$ & $0.022\pm0.000$ & $0.079\pm0.002$ & $0.069\pm0.000$ & $0.051\pm0.000$ \\
\textbf{}        & EVS${\uparrow}$ & $0.019\pm10.871$ & $0.911\pm0.056$ & $0.913\pm0.049$ & $0.545\pm0.122$ & $0.005\pm3.314$ & $0.801\pm0.005$ \\
\textbf{}        & MAPE${\downarrow}$ & $0.107\pm0.004$ & $0.028\pm0.000$ & $0.030\pm0.000$ & $0.136\pm0.005$ & $0.126\pm0.001$ & $0.009\pm0.000$ \\
\cmidrule(lr){1-8}
\multirow{5}{*}{\rotatebox[origin=c]{90}{\textbf{BART}}}        & RMSE${\downarrow}$ & $0.052\pm0.000$ & $0.041\pm0.001$ & $0.028\pm0.000$ & $0.107\pm0.003$ & $0.070\pm0.003$ & $0.068\pm0.001$ \\
\textbf{}        & $R^2$${\uparrow}$   & $0.856\pm0.014$ & $0.885\pm0.028$ & $0.995\pm0.001$ & $0.423\pm0.264$ & $0.743\pm0.222$ & $0.813\pm0.019$ \\
\textbf{}    & MAE${\downarrow}$  & $0.039\pm0.000$ & $0.028\pm0.000$ & $0.022\pm0.000$ & $0.074\pm0.003$ & $0.032\pm0.000$ & $0.036\pm0.000$ \\
\textbf{}        & EVS${\uparrow}$ & $0.875\pm0.009$ & $0.896\pm0.044$ & $0.960\pm0.001$ & $0.501\pm0.124$ & $0.747\pm0.213$ & $0.822\pm0.017$ \\
\textbf{}        & MAPE${\downarrow}$ & $0.070\pm0.002$ & $0.053\pm0.002$ & $0.036\pm0.000$ & $0.124\pm0.005$ & $0.063\pm0.001$ & $0.076\pm0.001$ \\
\bottomrule
\end{tabular}
\vspace{-5pt}
\caption{ASR results (mean$\pm$std) on different transformer-based models using Random Forest. Full results for Gradient Boosting (GB) and Linear Regression (LR) are presented in Table \ref{extend_table} (Appendix)}
\label{IP}
\vspace{-20pt}
\end{table}
\vspace{-5pt}
\section{Experiment Setup\protect\footnote{We refer the readers to the supplementary materials for implementation and reproducibility details.}}
\vspace{-5pt}
\noindent\textbf{Datasets.} We include 9 diverse publicly available classification corpus in the set $\pmb{\mathcal{D}}$, namely AG News~\cite{zhang2015character}, Amazon Reviews Full,  Amazon Reviews Polarity~\cite{keung-etal-2020-multilingual}, DBpedia~\cite{lehmann2015dbpedia}, Yahoo Answers, Yelp Reviews Full, Yelp Reviews Polarity~\cite{zhang2015character}, Banking77~\cite{casanueva-etal-2020-efficient}, and Tweet Eval Review~\cite{barbieri-etal-2020-tweeteval}

\noindent\textbf{Target Models.} We focus on studying the adversarial robustness of encoder-only transformer language models (LM) BERT~\cite{devlin-etal-2019-bert} and RoBERTa~\cite{liu2019RoBERTa}, which are often the standard baseline for text classification tasks. Moreover, we also report experiment results on decoder-only LM GPT2~\cite{radford2019language}, encoder-decoder LM BART~\cite{lewis2020bart} transformer models, and ELECTRA~\cite{clark2020electra}, which is an encoder-only model but trained with an additional discriminator.

\noindent\textbf{Interpolation and Extrapolation Evaluation.} For interpolation, we employ overlapped k-fold cross-validation of 80\%:20\% split and with $k{=}200$ to train and validate our framework on $\mathcal{\pmb{Q}}$. For extrapolation, data points are split based on their original dataset. For example, we have a list of datasets $\mathcal{D}_{l}$ and split them into three sets $\mathcal{D}_{1}$, $\mathcal{D}_{2}$, and $\mathcal{D}_{3}$ such that $\mathcal{D}_{l} = \bigcup \{\mathcal{D}_{1},\mathcal{D}_{2},\mathcal{D}_{3}\}$ and $\emptyset = \bigcap_{i,j \in \{1,2,3\} \textrm{ and } i\neq j} \{\mathcal{D}_{i},\mathcal{D}_{j}\}$ for training, validation and testing purposes, and to be more specific, $|\mathcal{D}_{1}|=5$, $|\mathcal{D}_{2}|=2$, and $|\mathcal{D}_{3}|=2$. The train, val, and test sets of the extrapolation prediction include the data points respectively sampled from datasets in $\mathcal{D}_{1}$, $\mathcal{D}_{2}$, and $\mathcal{D}_{3}$. With this strategy, the train, val, and test sets have different contexts and ranges which are useful for extrapolation testing purposes. 

\noindent\textbf{Evaluation Metrics.} We employ standard evaluation metrics of regression prediction problems including \textit{root mean square error} (RMSE), \textit{R squared} ($R^2$), \textit{mean absolute error and percentage error} (MAE, MAPE), \textit{explained variance score} (EVS). 

\begin{figure*}
\centering
\begin{subfigure}{.25\textwidth}
  \centering
  \includegraphics[width=\textwidth]{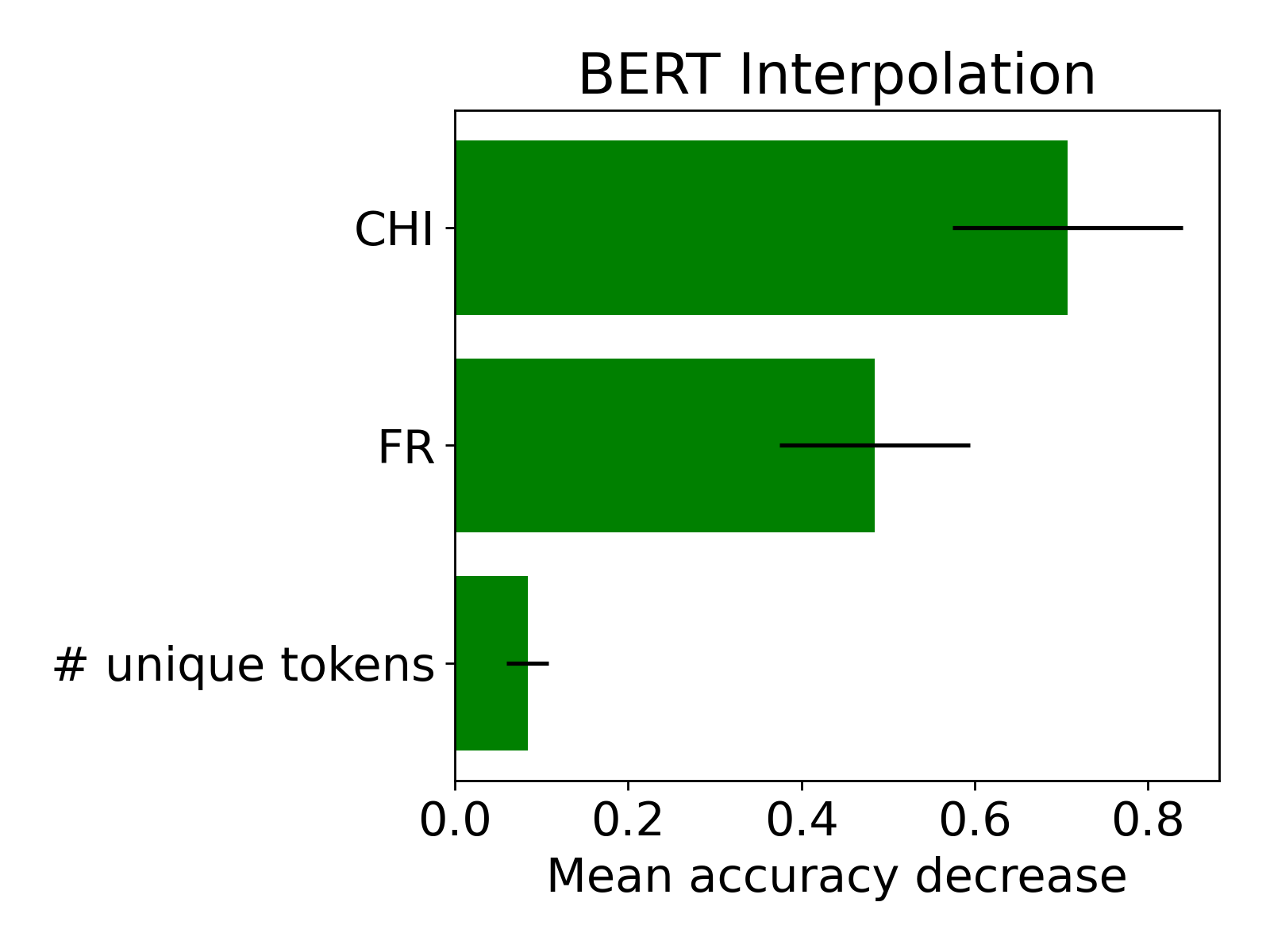}
  \caption{\textbf{Interpolation}-\textbf{BERT}}
  \label{inter_bert_fpi}
\end{subfigure}\hfill%
\begin{subfigure}{.25\textwidth}
  \centering
  \includegraphics[width=\textwidth]{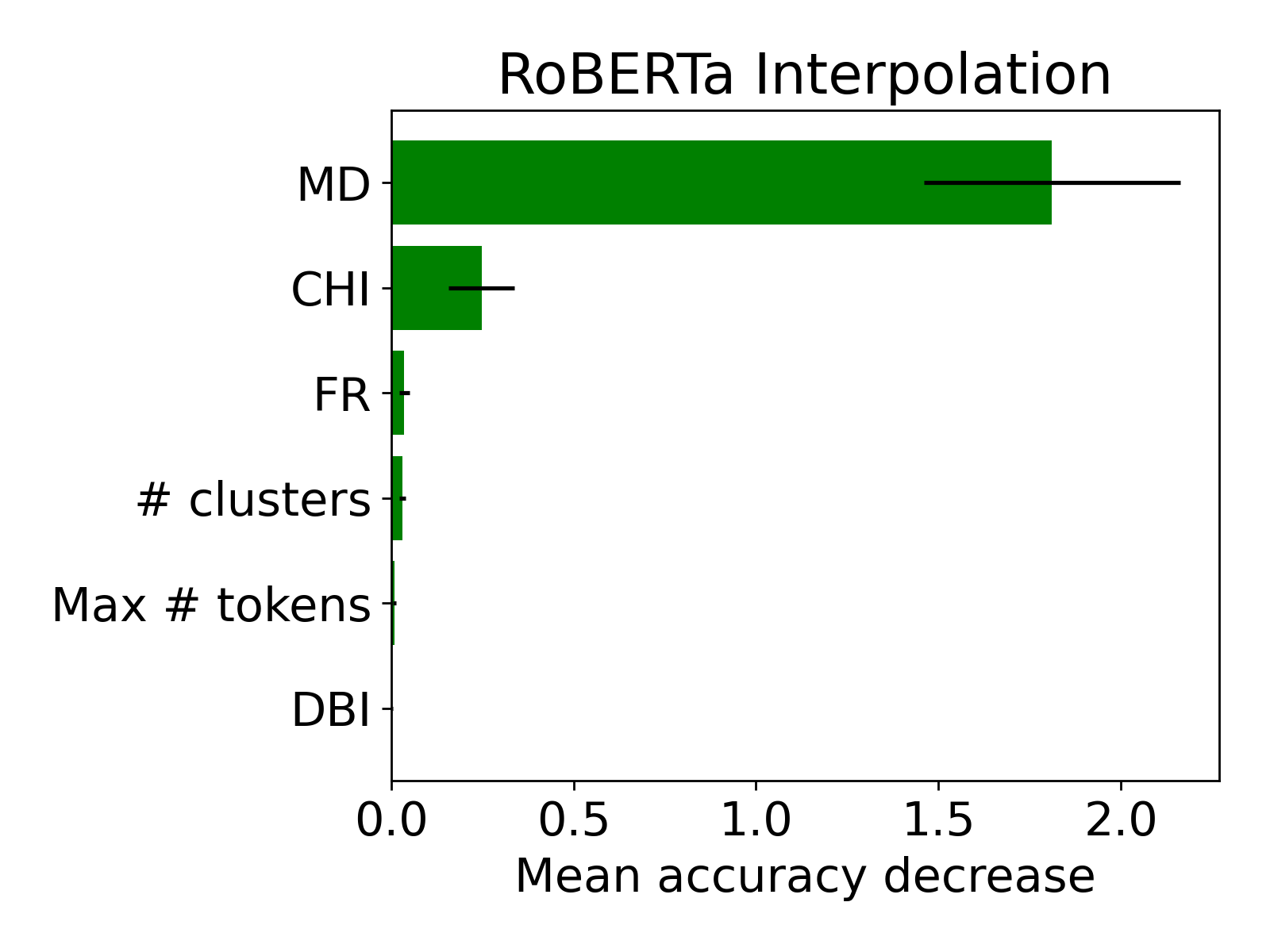}
  \caption{\textbf{Interpolation}-\textbf{RoBERTa}}
  \label{inter_roberta_fpi}
\end{subfigure}\hfill%
\begin{subfigure}{.25\textwidth}
  \centering
  \includegraphics[width=\textwidth]{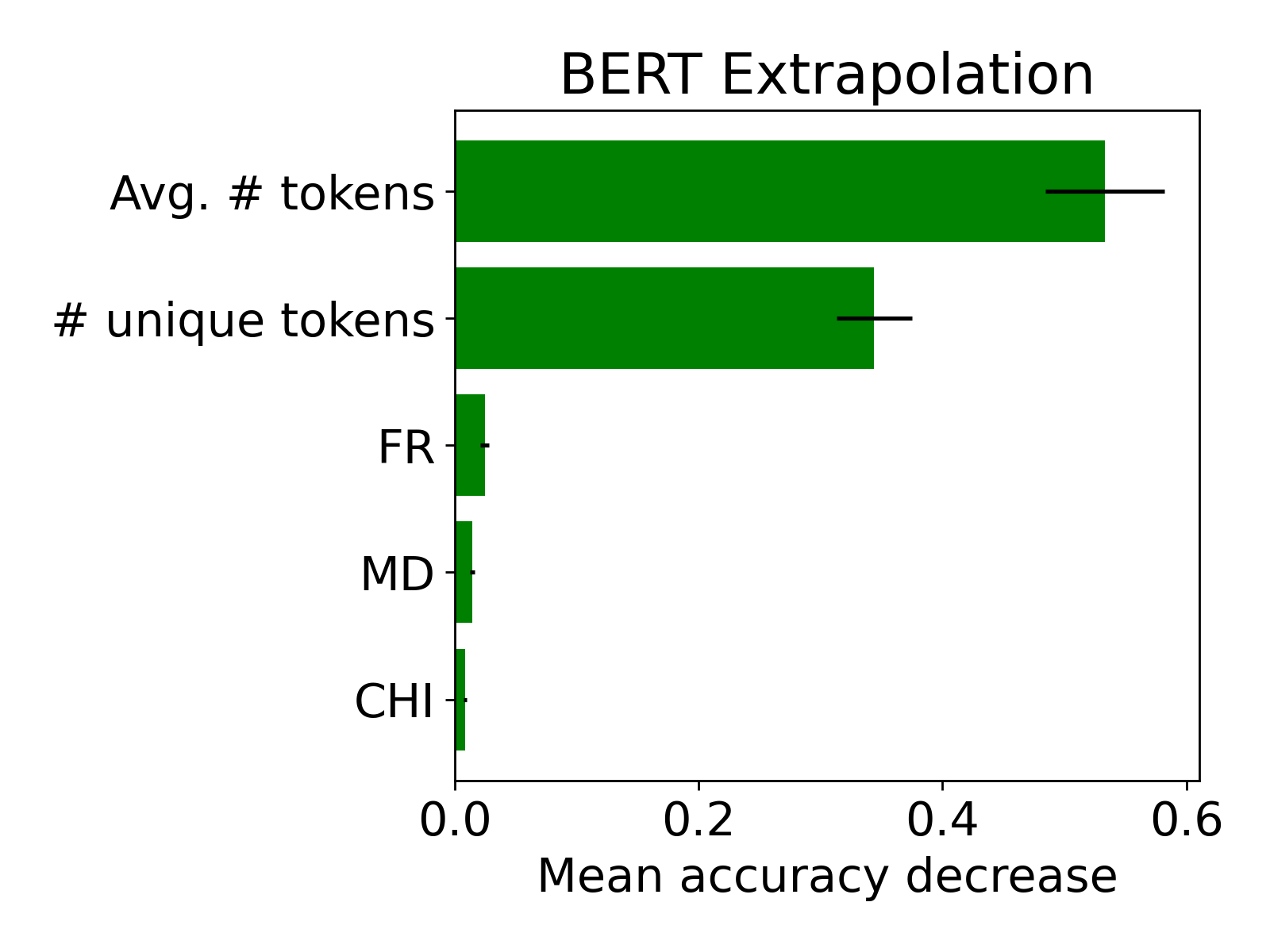}
  \caption{\textbf{Extrapolation}-\textbf{BERT}}
  \label{extra_bert_fpi}
\end{subfigure}\hfill%
\begin{subfigure}{.25\textwidth}
  \centering
  \includegraphics[width=\textwidth]{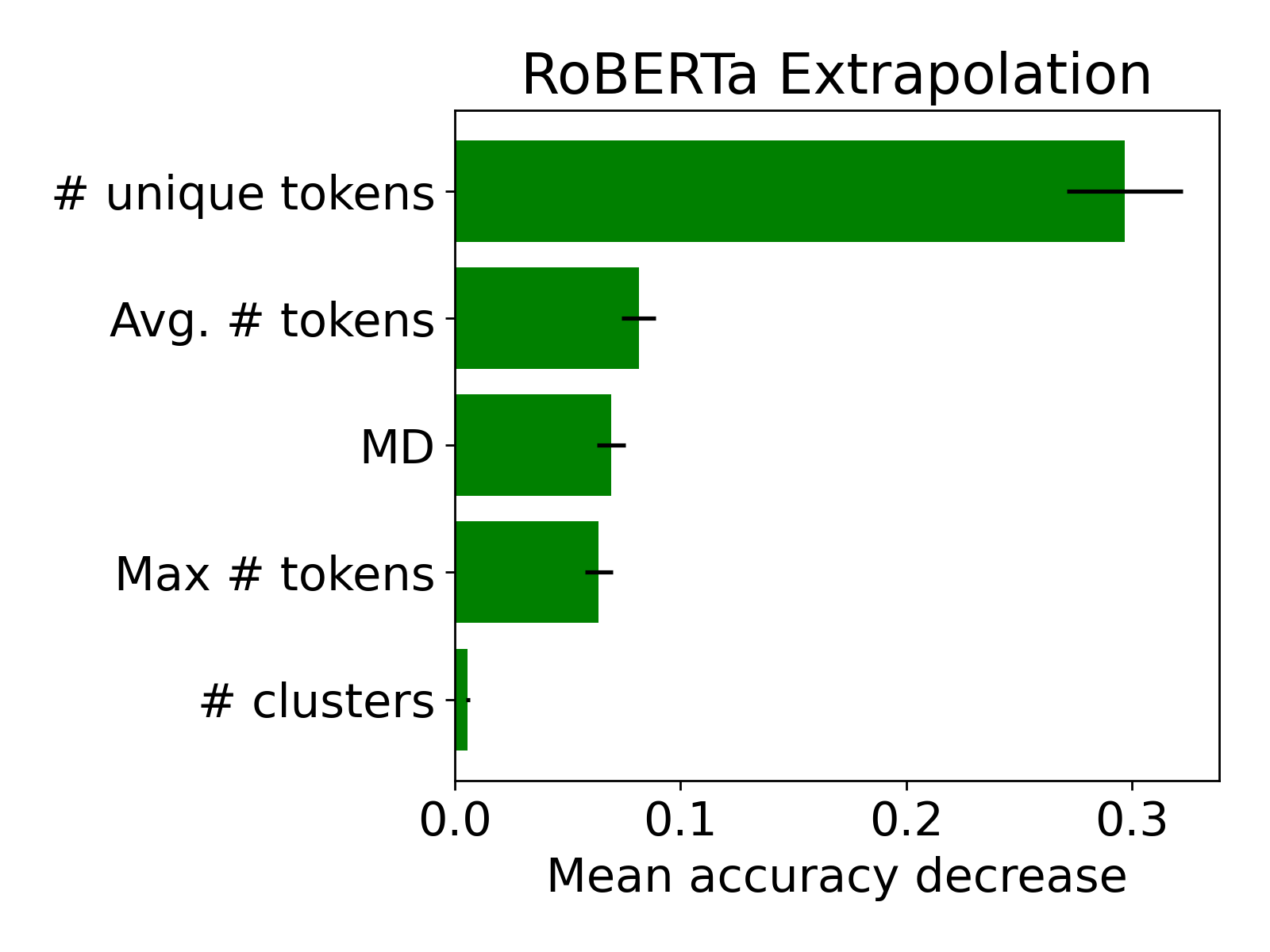}
  \caption{\textbf{Extrapolation}-\textbf{RoBERTa}}
  \label{extra_roberta_fpi}
\end{subfigure}
\caption{Importance of the best Random Forest regression model's most important features in predicting ASRs of BERT and RoBERTa in interpolation and extrapolation setting. 
\vspace{-0.1in}
}
\label{feature_ranking}
\end{figure*}

\vspace{-5pt}
\section{Results, Analyses, and Discussions\protect\footnote{We refer the readers to Section \ref{discussion} for more details.}}
\label{results}
\vspace{-5pt}
\noindent \textbf{\textit{Finding 1: Fine-tuning data have a strong correlation with Model Robustness.}}
Table \ref{IP} shows the results of ASRs under both interpolation and extrapolation settings. Random Forest predictor achieves the best results, followed by Gradient Boosting and Linear Regression in most cases except for extrapolation prediction on RoBERTa. Regarding to MAE, Random Forest scores are as low as 0.025 and 0.037 for interpolation prediction on BERT and RoBERTa. It also achieves reasonable extrapolation prediction with MAE of only around 0.045 and 0.044 on BERT and RoBERTa. Requiring only \textit{one initial training}, our framework shows to be effective at benchmarking the adversarial robustness of BERT and RoBERTa with only a lightweight Random Forest predictor. 


\noindent \textbf{\textit{Finding 2: Embedding distribution and token-based statistics features are among the most influential indicators of adversarial robustness.}} Finding 1 demonstrates that our engineered features are highly informative about the fine-tuned model's robustness. Fig. \ref{feature_ranking} further summarizes the order of influence of each feature in the case of Random Forest, which is the best regression predictor we found in Finding 1. We only show features that have an average influence score twice greater than their variance. 

Overall, embedding distribution and token-based statistics are the two groups of most influential features. In interpolation, CHI, FR, and \# of unique tokens have a significant influence on the adversarial robustness of BERT (Fig. \ref{inter_bert_fpi}), whereas such feature set of RoBERTa also includes MD (Fig. \ref{inter_roberta_fpi}). We also observe a similar pattern in predicting the adversarial robustness of BERT (Fig. \ref{extra_bert_fpi}) and RoBERTa (Fig. \ref{extra_roberta_fpi}) in extrapolation prediction.

\begin{figure*}[!htb]
\centering
\begin{subfigure}{.25\textwidth}
  \centering
  \includegraphics[width=\textwidth]{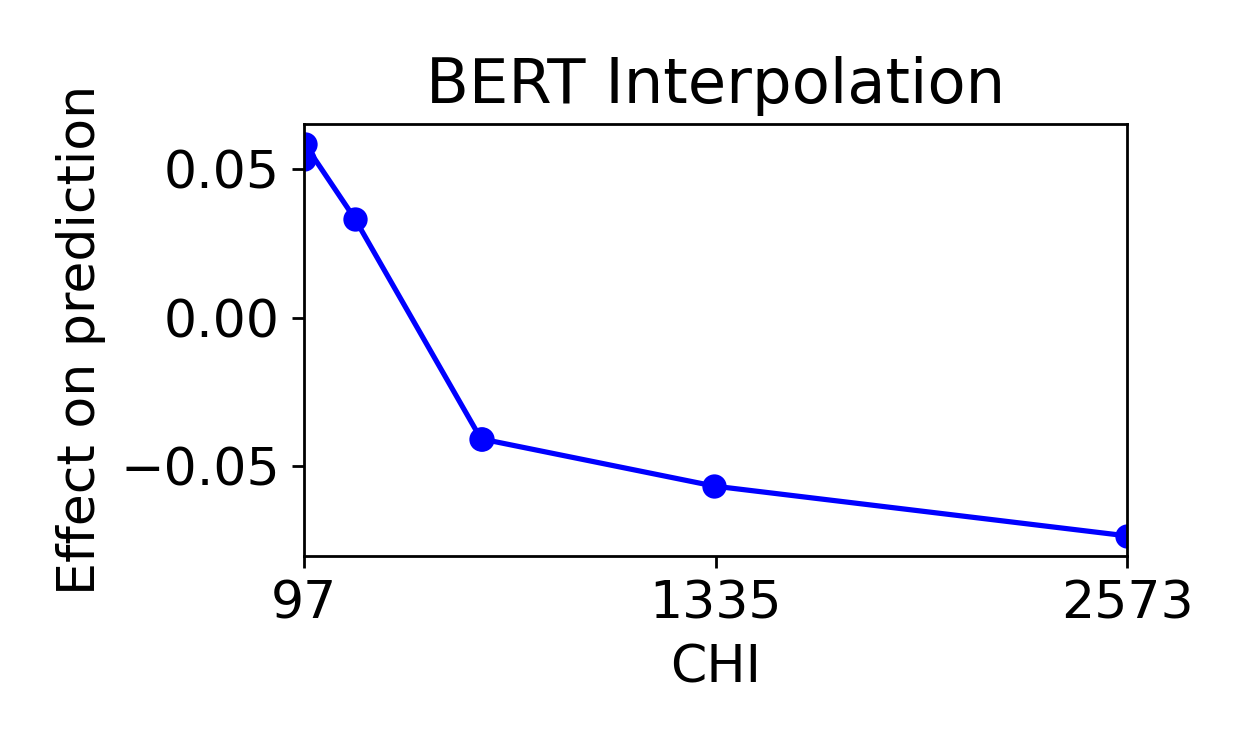}
\end{subfigure}%
\begin{subfigure}{.25\textwidth}
  \centering
  \includegraphics[width=\textwidth]{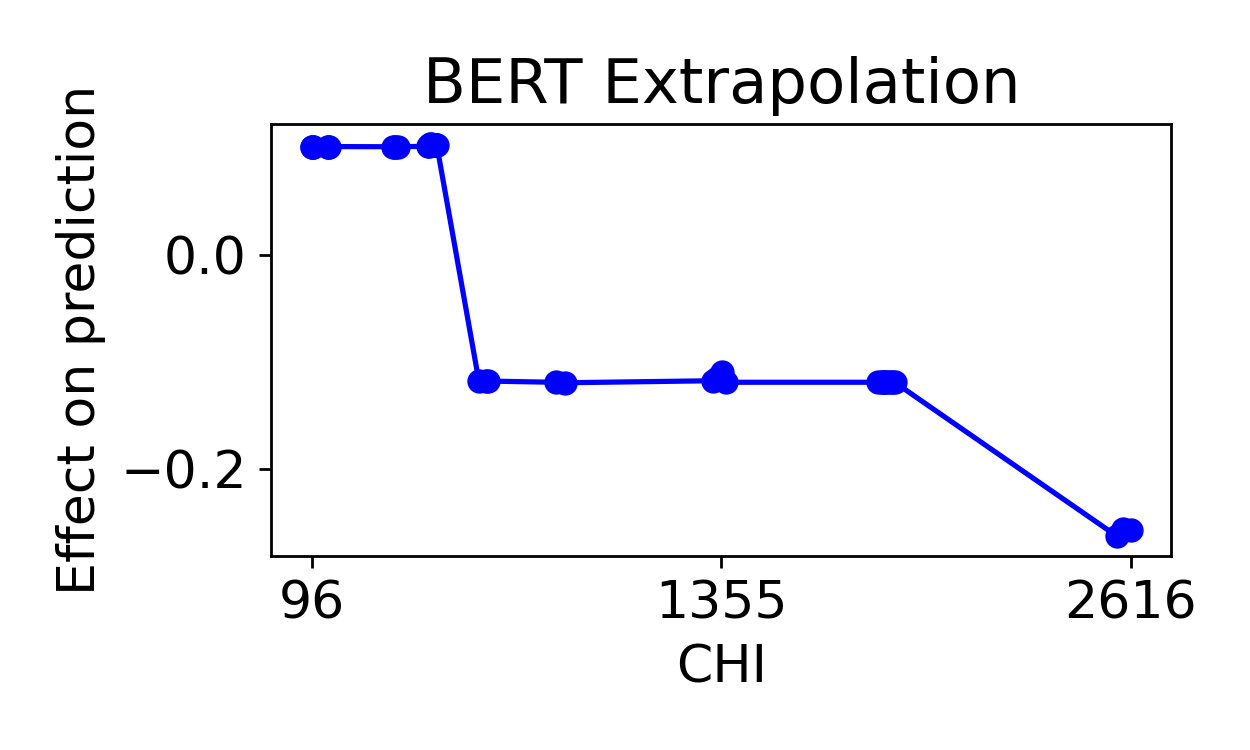}
\end{subfigure}%
\begin{subfigure}{.25\textwidth}
  \centering
  \includegraphics[width=\textwidth]{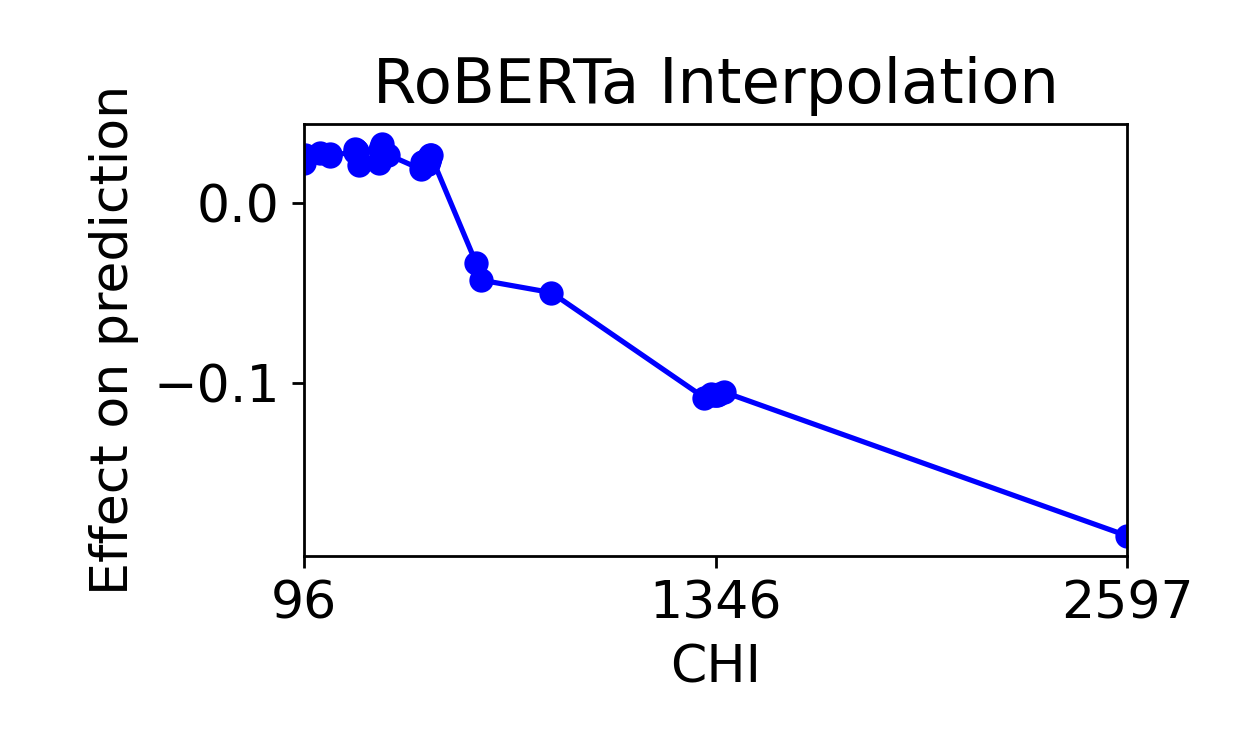}
\end{subfigure}%
\begin{subfigure}{.25\textwidth}
  \centering
  \includegraphics[width=\textwidth]{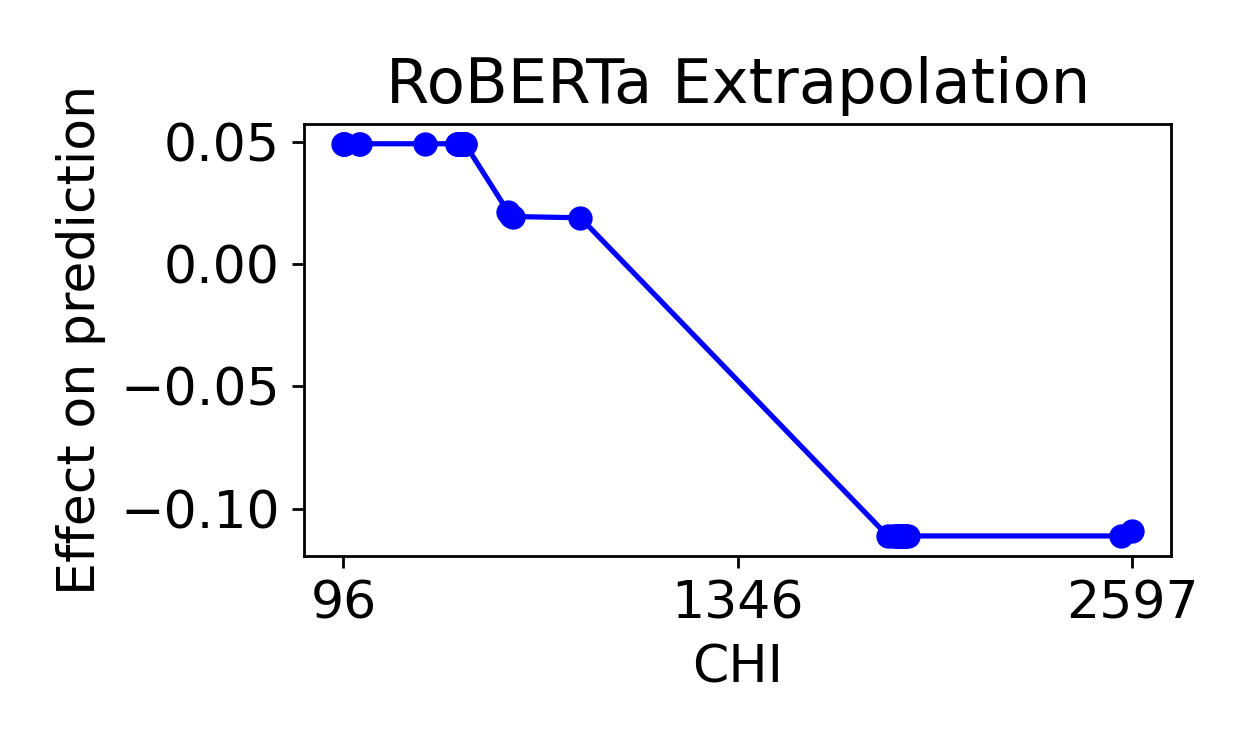}
\end{subfigure}
\begin{subfigure}{.25\textwidth}
  \centering
  \includegraphics[width=\textwidth]{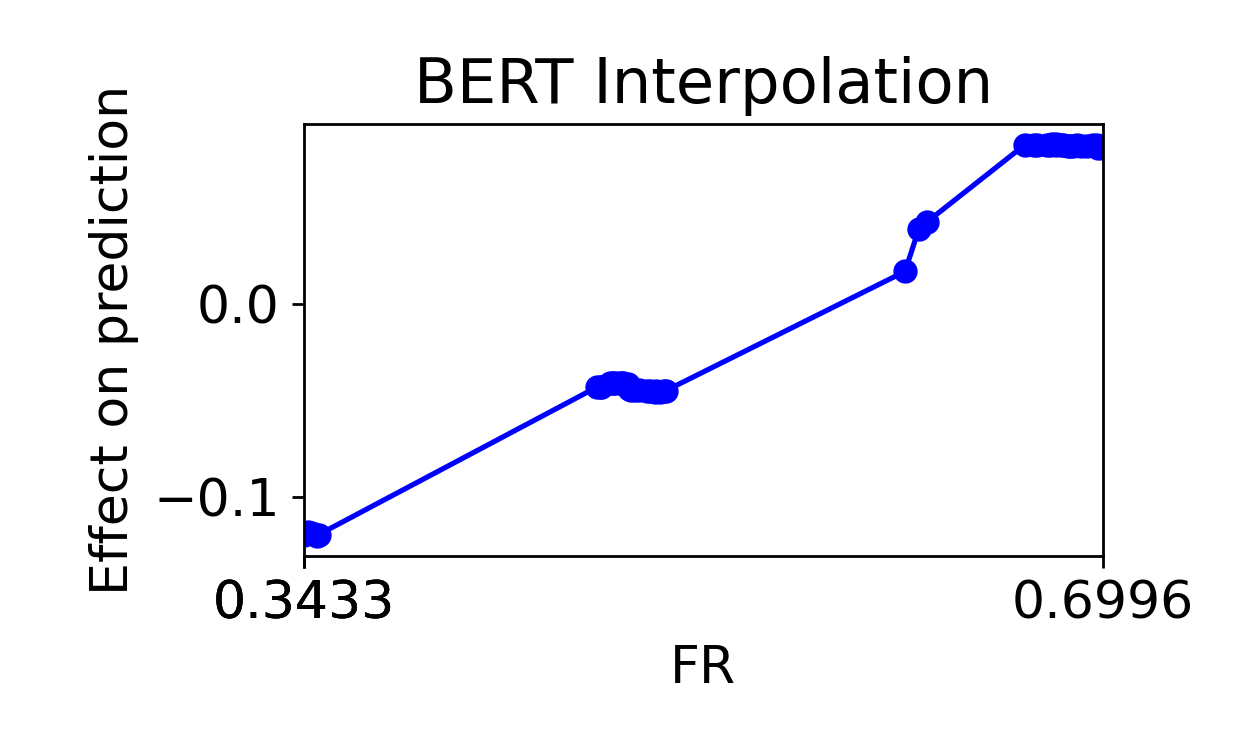}
\end{subfigure}%
\begin{subfigure}{.25\textwidth}
  \centering
  \includegraphics[width=\textwidth]{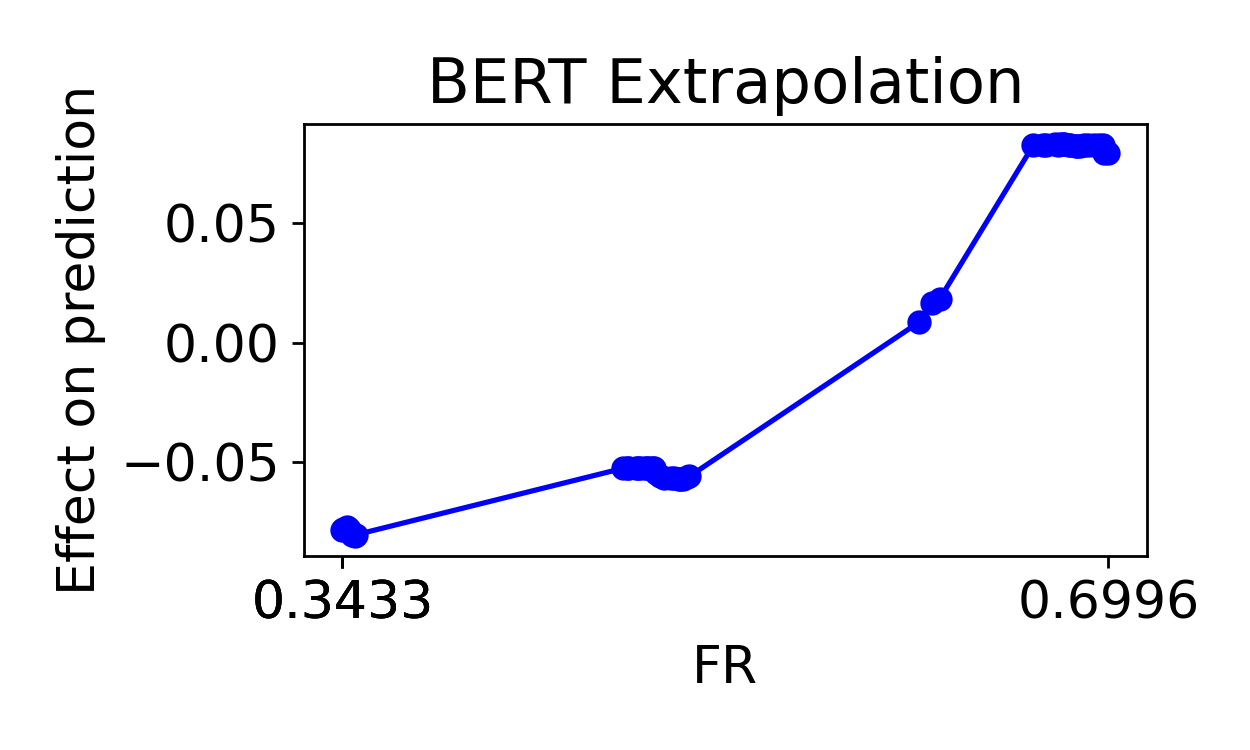}
\end{subfigure}%
\begin{subfigure}{.25\textwidth}
  \centering
  \includegraphics[width=\textwidth]{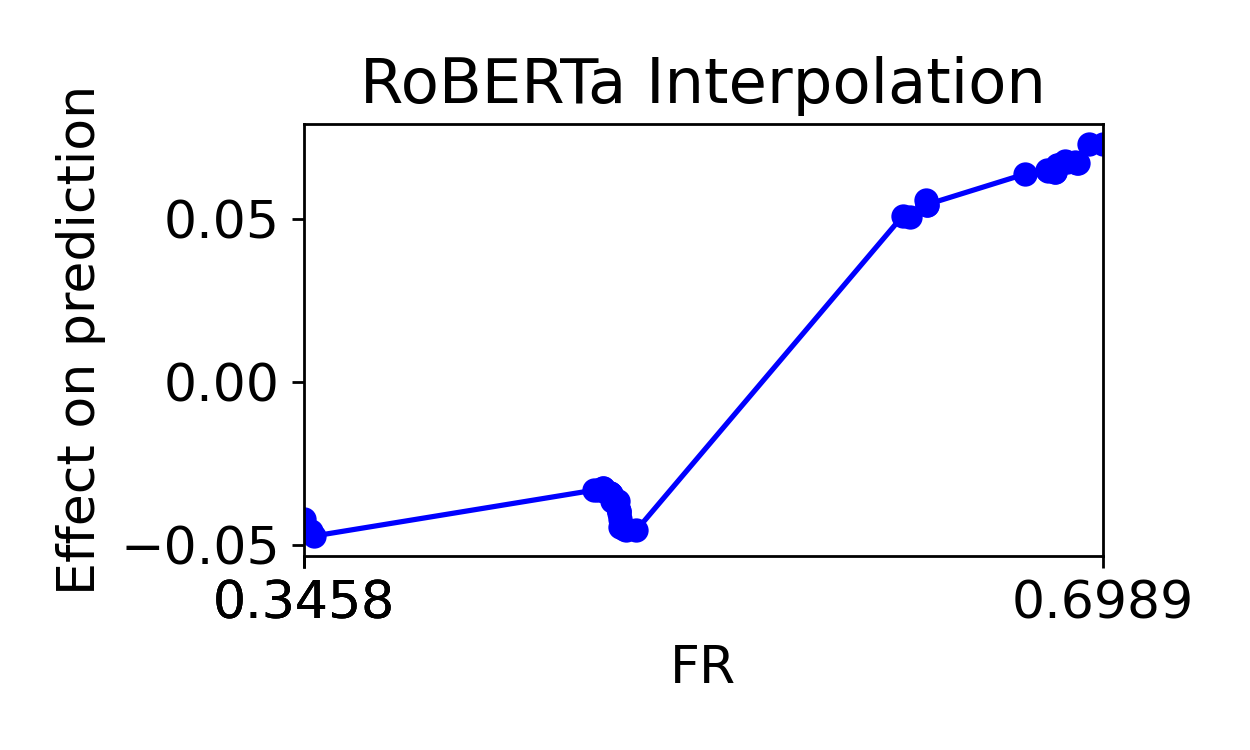}
\end{subfigure}%
\begin{subfigure}{.25\textwidth}
  \centering
  \includegraphics[width=\textwidth]{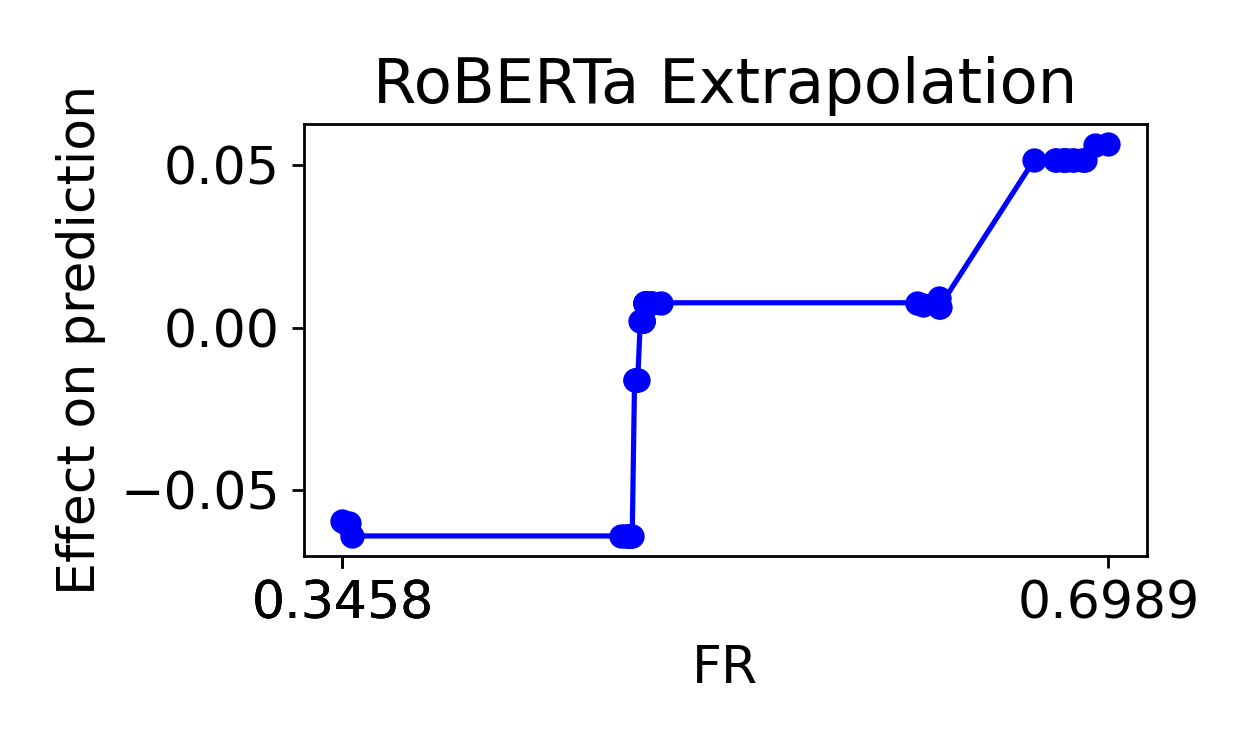}
\end{subfigure}
\begin{subfigure}{.25\textwidth}
  \centering
  \includegraphics[width=\textwidth]{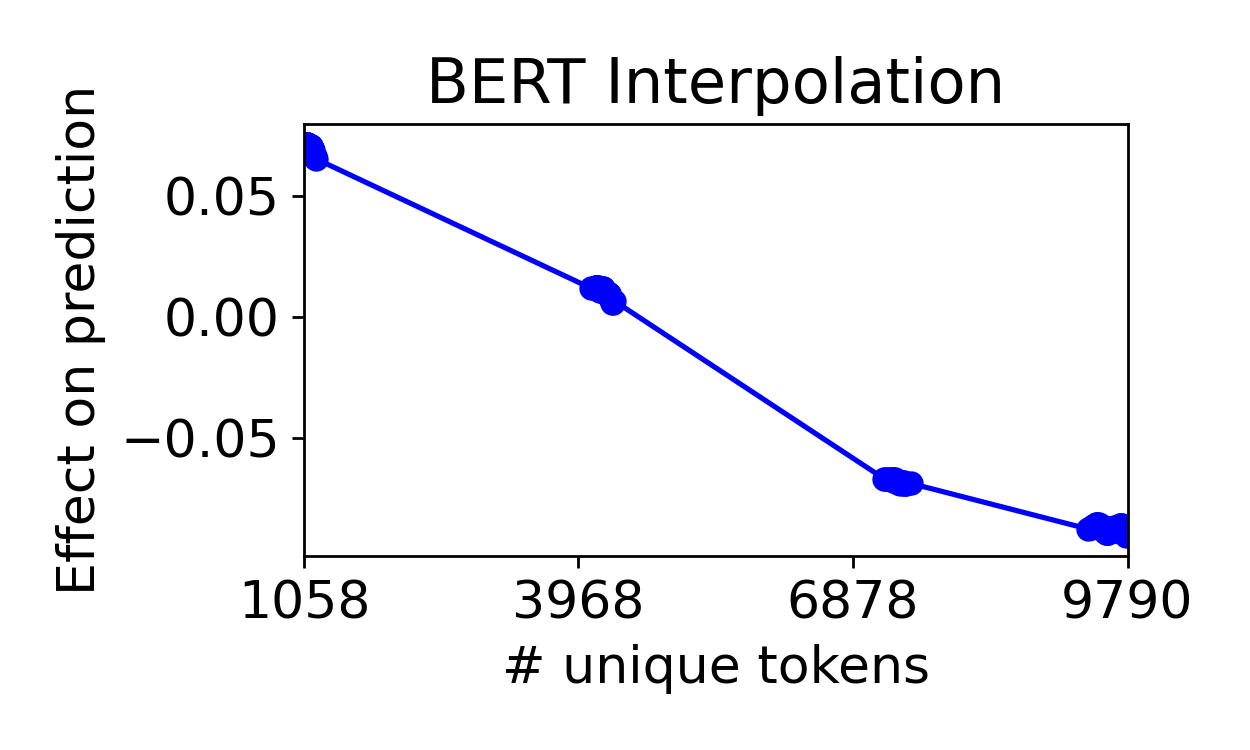}
\end{subfigure}%
\begin{subfigure}{.25\textwidth}
  \centering
  \includegraphics[width=\textwidth]{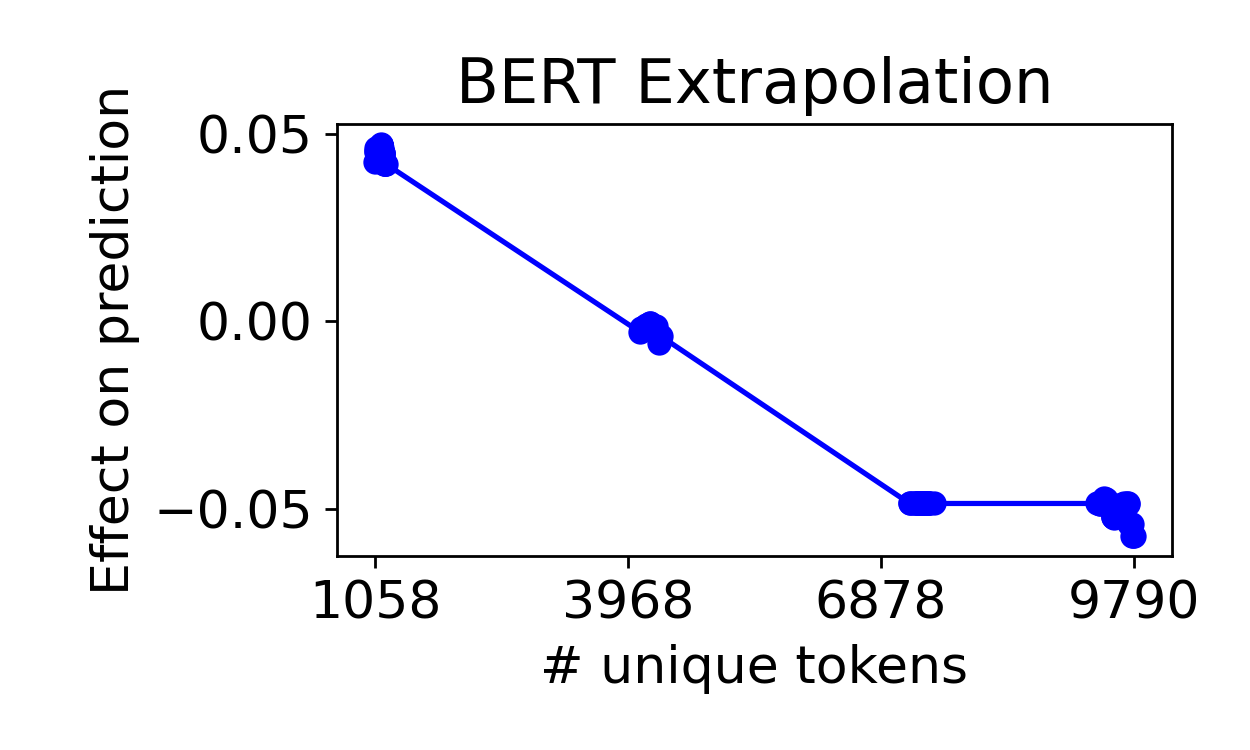}
\end{subfigure}%
\begin{subfigure}{.25\textwidth}
  \centering
  \includegraphics[width=\textwidth]{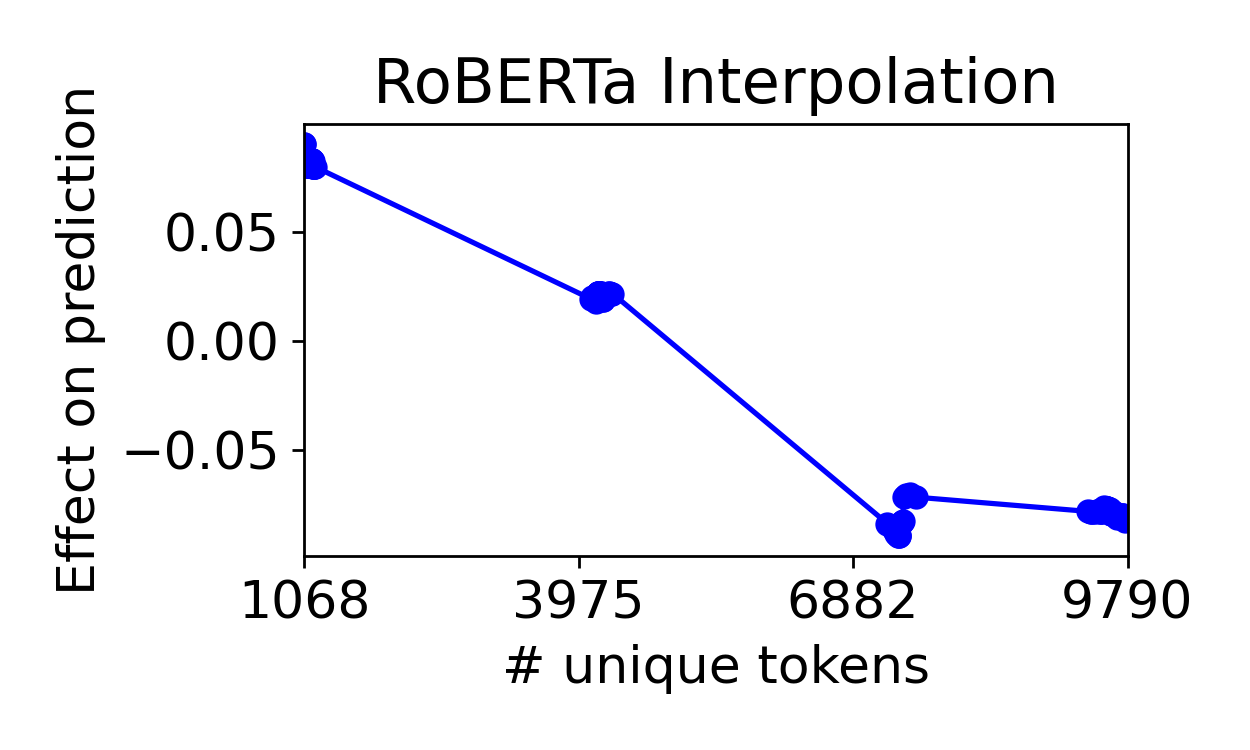}
\end{subfigure}%
\begin{subfigure}{.25\textwidth}
  \centering
  \includegraphics[width=\textwidth]{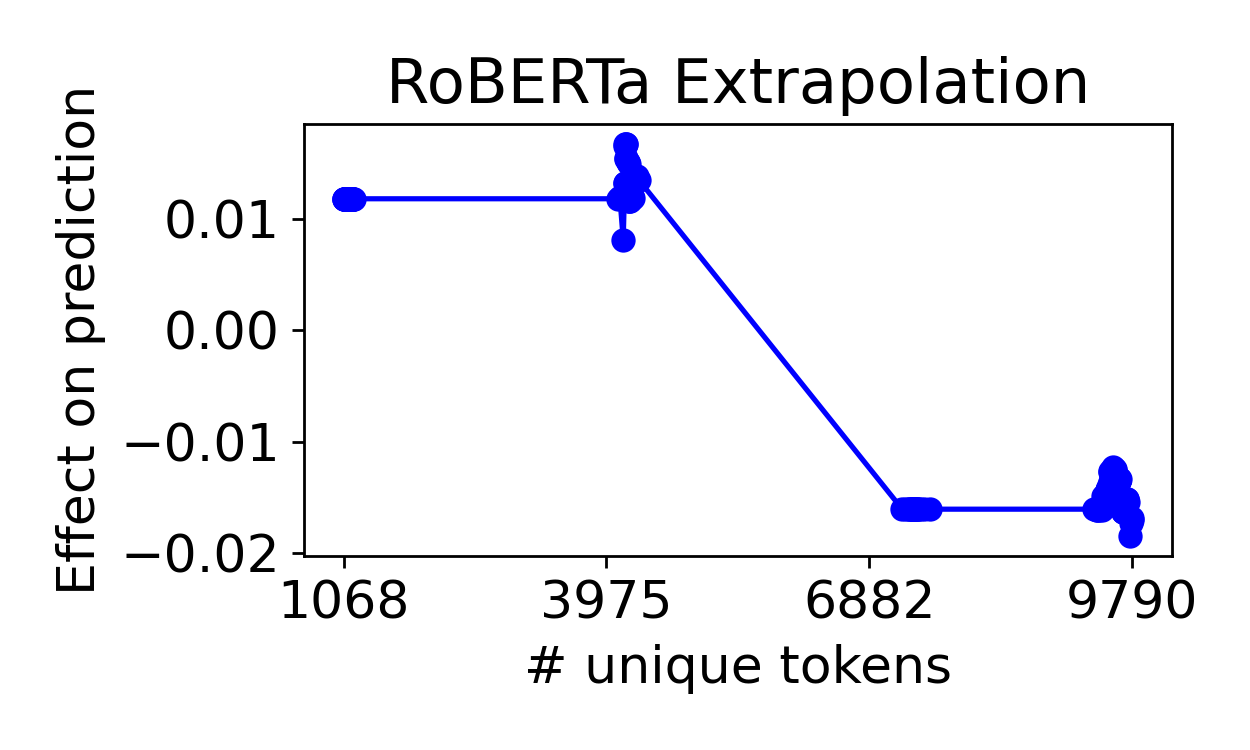}
\end{subfigure}
\begin{subfigure}{.25\textwidth}
  \centering
  \includegraphics[width=\textwidth]{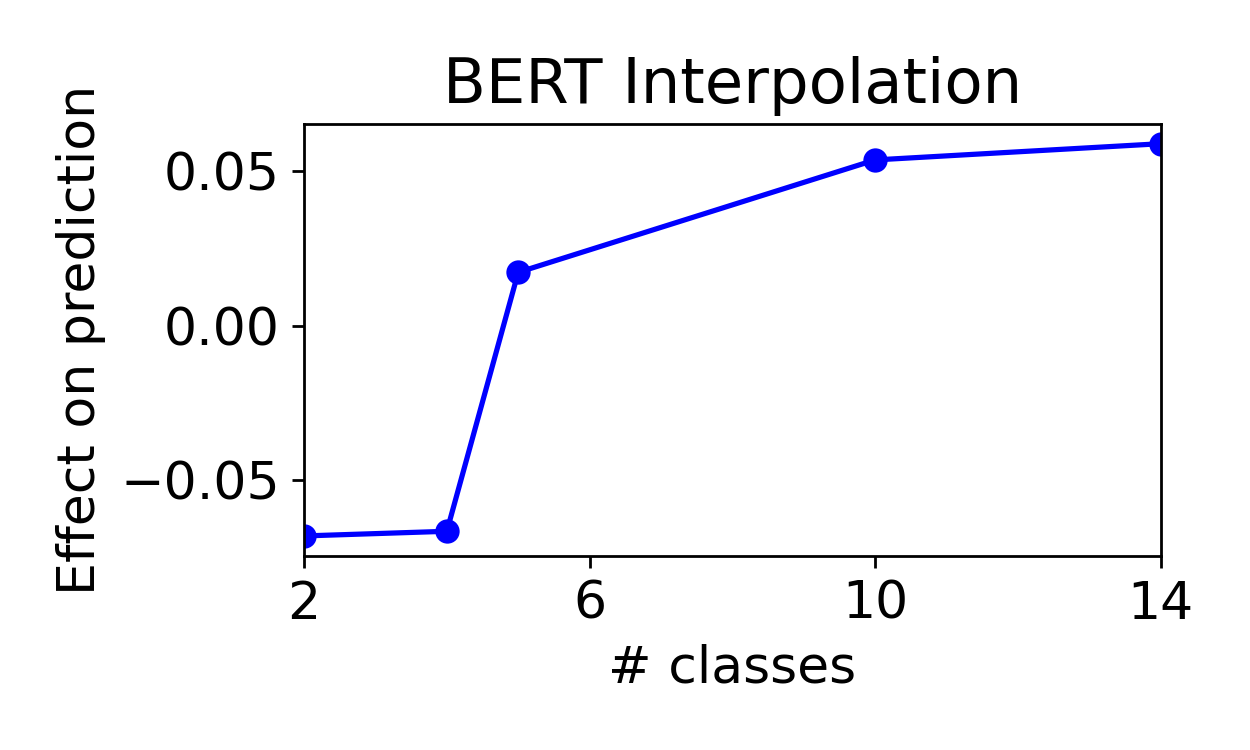}
\end{subfigure}%
\begin{subfigure}{.25\textwidth}
  \centering
  \includegraphics[width=\textwidth]{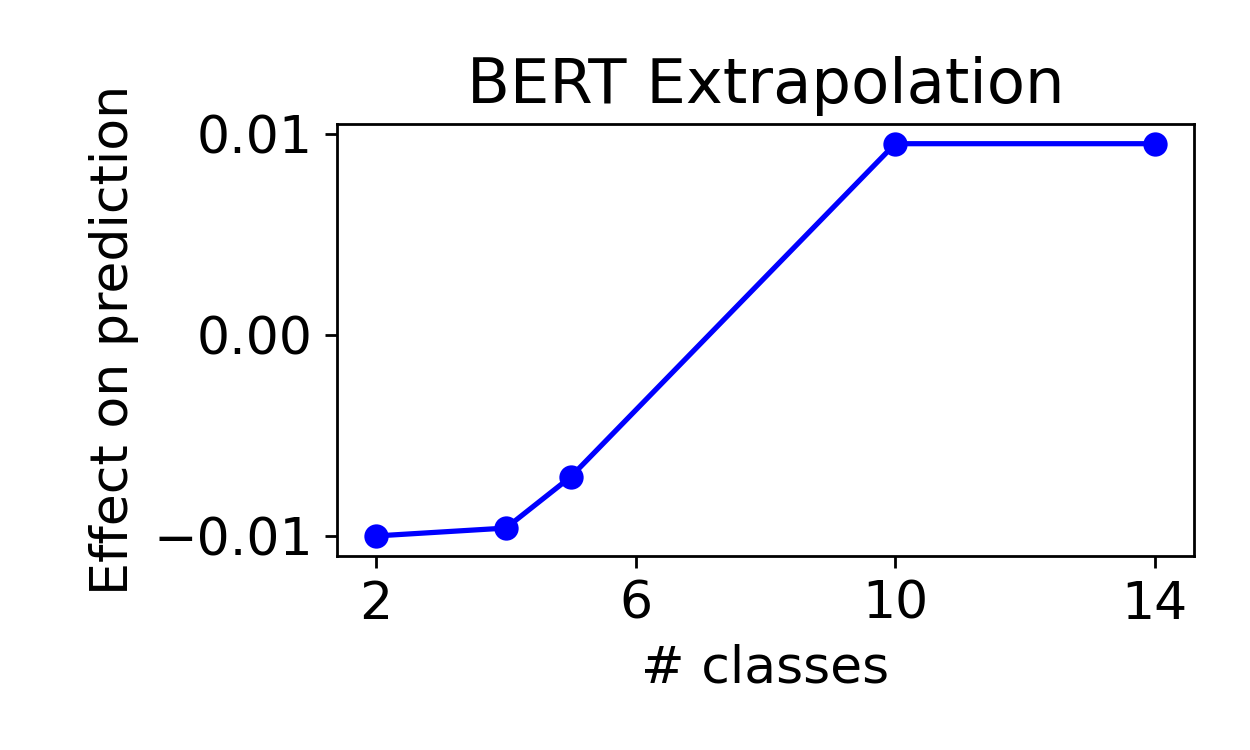}
\end{subfigure}%
\begin{subfigure}{.25\textwidth}
  \centering
  \includegraphics[width=\textwidth]{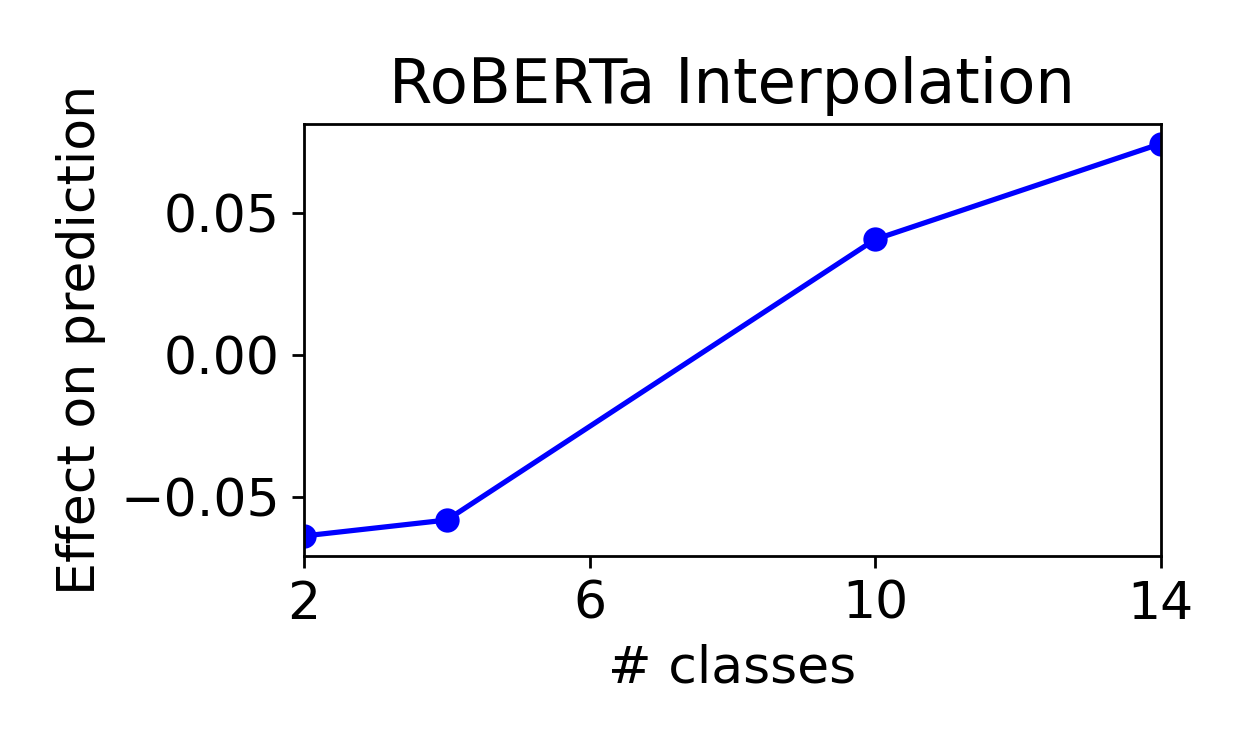}
\end{subfigure}%
\begin{subfigure}{.25\textwidth}
  \centering
  \includegraphics[width=\textwidth]{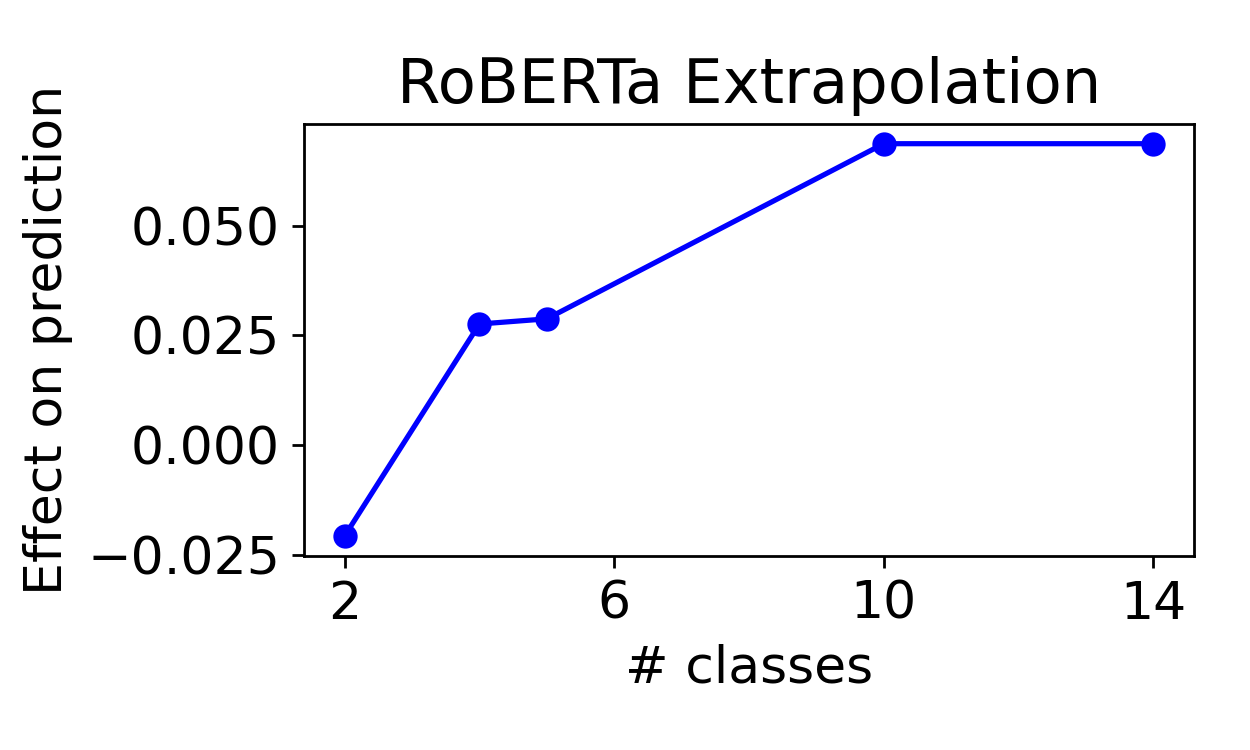}
\end{subfigure}
\caption{CHI, FR, \# of tokens, and \# of classes (top to bottom) show clear correlation patterns with ASRs. 
}
\vspace{-15pt}
\label{key_ft_analyze}
\end{figure*}

\noindent \textbf{\textit{Finding 3: CHI, FR, \# unique tokens and \# classes have clear correlations with ASR.}}
Fig. \ref{key_ft_analyze} provides the correlation between notable features discussed in Finding 2 and how they influence the ASR prediction on average. These results show that the distances among classes in the embedding space--i.e., class separation sub-group (Fig. \ref{taxanomy}), are highly indicative of the adversarial robustness of the fine-tuned models. When the embedding among classes disperses in the space and is not concentrated, FR feature has a low value and CHI feature has a high value, which correlates to a greater robustness against adversarial examples. The opposite also holds as well, as illustrated in Fig. \ref{embed}.
Furthermore, token-based statistics of the dataset such as \# of unique tokens and \# of classes also contribute to the influence on adversarial robustness. As \# of classes increases, the embedding space becomes denser and clusters among prediction labels show more overlaps, the less robustness observed in the fine-tuned models. Moreover, a large \# of unique tokens often informs a diverse fine-tuning dataset, which makes the pre-trained transformer-based models more generalizable and hence more difficult to attack.

\noindent \textbf{\textit{Error Analysis.~\footnote{We refer the readers to experiment setup for error analysis in supplementary materials.}}} The top three error-inducing features in ASR prediction are DBI, \# of classes, and MR. Unlike MD, FDR, and CHI, DBI lacks robustness and fails to accurately represent embedding concentration because it is based on the distance to the nearest cluster compared to the original cluster. The increasing \# of classes makes the decision boundary more complicated and greatly affects ASR, but when a saturation threshold is crossed, this phenomenon no longer occurs. This explains our observation that although there is a strong correlation between the \# of classes and ASR shown in Finding 3, our predictor has poor performance when the \# of classes is greater than 10. CNN calculates the misclassification rate (MR) of the surrogate model, leveraging its focus on local structures, whereas the transformer model relies on global dependencies. Consequently, in some cases, while the MR of CNN may vary significantly, the transformer's adversarial robustness remains relatively consistent.
\begin{figure}[tb]
    \centering
    \includegraphics[width=\linewidth]{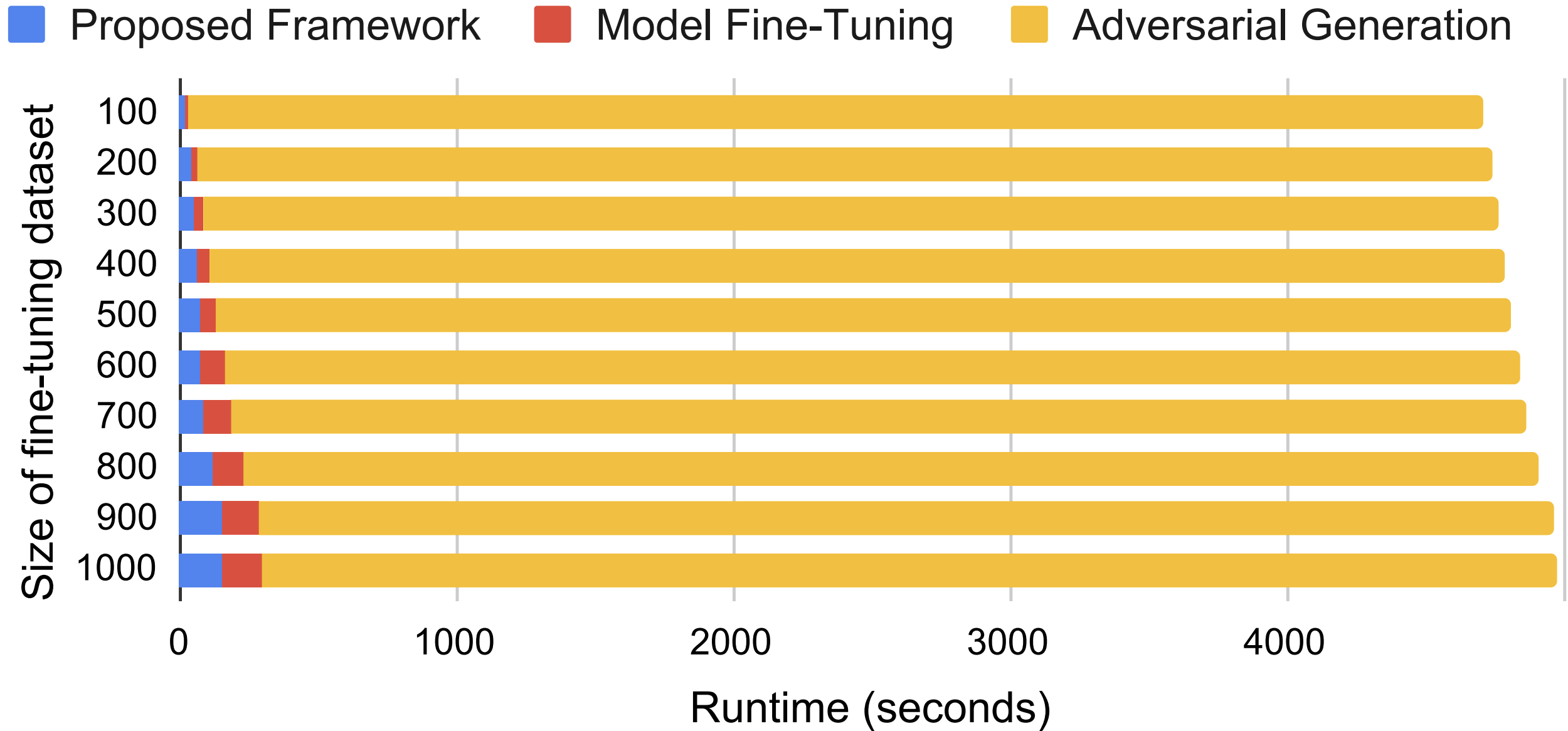}
    \caption{Our framework significantly improve running time, be it \textit{30}$\times$ to \textit{193}$\times$ faster than traditional methods with \textit{Model Fine-Tuning}+\textit{Adversarial Generation} steps.}
    \label{runtime}
    \vspace{-0.1in}
\end{figure}
\begin{table}[tb]
\centering
\resizebox{\columnwidth}{!}{%
\begin{tabular}{lcccc}
\toprule
\textbf{METRIC} & \textbf{BERT} & \textbf{Distil-BERT} & \textbf{RoBERTa} & \textbf{Distil-RoBERTa} \\ \hline
RMSE${\downarrow}$ & 0.070          & 0.100 & \textbf{0.061} & 0.072          \\
$R^2$${\uparrow}$   & \textbf{0.806} & 0.621 & 0.782          & 0.740          \\
MAE${\downarrow}$  & \textbf{0.045} & 0.075 & 0.052          & 0.049          \\
EVS${\uparrow}$  & 0.812          & 0.790 & \textbf{0.918} & 0.760          \\
MAPE${\downarrow}$ & 0.145          & 0.173 & 0.139          & \textbf{0.109} \\ \bottomrule
\end{tabular}%
}
\caption{We train on the robustness of 3 models and test on the remaining one to test the transferability between transformer models of robustness predictor. The top row indicates the model to be tested.}
\label{transfer_tab}
\vspace{-18pt}
\end{table}

\vspace{-5pt}
\section{Another Tool for Robustness Analysis}
\vspace{-5pt}
\noindent \textbf{\textit{The proposed approach saves significant runtime in estimating adversarial robustness with reasonable accuracy.}} The advantage of our method lies in skipping the adversarial example generation of four attacking methods used for evaluating adversarial robustness, making our inference time 30x--193x faster than the traditional approach when evaluating adversarial robustness on 100 examples (Fig. \ref{runtime}). For example, when inferring the robustness of a transformer model fine-tuned on 900 test samples shown in Fig. \ref{runtime}, our method takes 153.02s including feature extraction (152.48s) and robustness inference by Random Forest (0.18s). Conversely, the traditional method takes 4807.83s including Fine-tuning PLM (130.51s) and Adversarial Generation (4677.32s). As a result, our proposed framework is 31.4 times faster.
 
Thanks to the accurate predictions discussed in Finding 1 of Section \ref{results} and fast runtime speed, our framework can be used as a \textit{additional tool} for quickly pinpointing adversarial robustness.

\noindent \textbf{\textit{Generalization between transformer-based text classifiers.}}
We perform robustness predictor training on 3 models and test on the remaining one. The results of Table \ref{transfer_tab} show that R2 and RMSE range from 0.62-0.81 and 0.06-0.10, respectively. This indicates the transferability between transformer-based text classifiers of our robustness predictor.

\noindent \textbf{\textit{Support adversarial training.}}
\begin{figure}[tb]
    \centering
    \includegraphics[width=\linewidth]{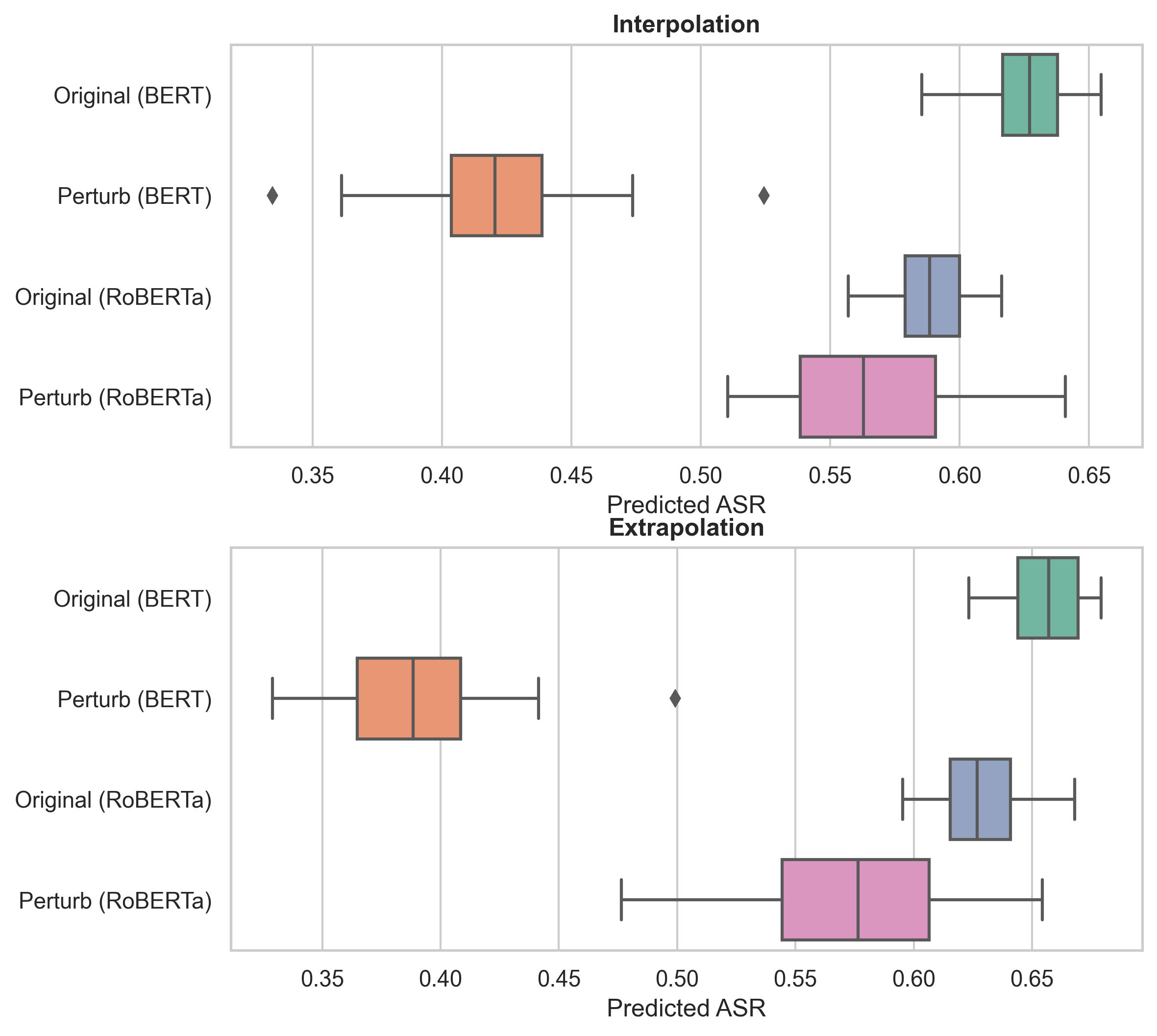}
    \vspace{-10pt}
    \caption{ASR Prediction for BERT and RoBERTa with and without adversarial training in both interpolation and extrapolation.
    }\label{cp_method}
    \vspace{-18pt}
\end{figure}
We perform adversarial robustness prediction ability of the best performing Random Forest predictor in the case of adversarial training. Specifically, we predict the robustness of BERT and RoBERTa on a fine-tuning dataset that includes both original and perturbed texts. 
Fig. \ref{cp_method} summarizes the results. Our Random Forest framework consistently outputs lower ASRs, thus informing more robust BERT and RoBERTa models under both interpolation and extrapolation. This shows that our engineered features can capture nuanced changes in the text embedding space of the fine-tuning datasets and inform the Random Forest predictor to respond accordingly even without observing any adversarial examples during training. 

\noindent \textbf{\textit{Robustness to statistical randomness.}}
Because ASR is a statistical metric; inevitably, the robustness predictor itself is not robust to randomness. 
Evaluations for the prediction of Random Forest in Table \ref{IP} also show its consistency in that results just vary from 0.00-0.01 and 0.00-0.03 in interpolation and extrapolation settings.
\vspace{-5pt}
\section{Further Discussion}
\label{discussion}
\vspace{-5pt}

\subsection{Confounding Factors}
\label{confounding}
\vspace{-5pt}
This work assumes two main factors affecting adversarial robustness: training data and a model’s architecture. Most existing works analyze a model’s adversarial robustness after it has been fine-tuned on the training data, thus mixing the two factors, making the analysis of how much training data (alone) correlates with adversarial robustness difficult \cite{xu2019understanding,zheng-etal-2023-characterizing}. Therefore, in this work, we separate the two factors and focus only on using training data as the input to measure the adversarial robustness for a specific model’s architecture without fine-tuning such a model (fixed second variable–i.e., model architecture, and vary the first variable–i.e., training data).

Another possible confounding factor that affects both the training data and a model’s adversarial robustness is the label and data curation process. For example, some malicious actors might intentionally poison the training data to affect the training features, labels, and a model’s adversarial robustness. We assume that all of our training data is clean (we use official, published sources of all datasets) and we do not expect any confounding factors might affect our analysis.

There might be other confounding factors that we might overlook. Within the pioneering nature of our work on this research topic, we hope to see future works that explore those confounding factors, for example, from the perspective of causality.

\vspace{-5pt}
\subsection{Contextual Features}
\vspace{-5pt}
More nuanced, contextual, and semantic features of the training data will be useful for adversarial robustness prediction. In fact, our proposed framework already leverages the more nuanced aspects of fine-tuning data such as context and semantics by representing the original text using the Universal Sentence Encoder (USE) during the feature engineering step in Sec. \ref{extract_features}. However, there might be other more complex features that would only be captured using complex neural network models, which then might inhibit interpretability and increase the runtime. For instance, if we aim to duplicate our training process but opt for an efficient neural network like CNN, we would have to train it with training examples. Each of them would consist of 900 sentences, where the average sentence length is 60 tokens. This results in an input size as large as $900\times256$ for each training example, with 256 as the word embedding size. This can be also considered as an image of $900\times256$ dimension, which is much larger compared to images of $28\times28$ dimension in the MNIST dataset. If we are not using CNN but a simple 1-hidden layer neural network with only 10 neuron units, we would need over 138M model parameters. We would expect a much longer runtime compared to our approach.

\vspace{-5pt}
\subsection{Relationship between compactness of input embedding and adversarial robustness}
\vspace{-5pt}
\cite{pal2024adversarial} introduces the theoretical concept of input data concentration, demonstrating that a robust classifier emerges when the input embedding is concentrated. On the other hand, \cite{si-etal-2021-better} employs augmentation techniques to enhance the density and concentration of the input embedding. Consequently, classifiers trained on such embeddings exhibit greater robustness. However, the approach of \cite{si-etal-2021-better} lacks an explicit explanation of the underlying rationale and fails to establish a direct correlation between input embedding concentration and adversarial robustness. Our research complements the findings of \cite{si-etal-2021-better} by revealing that sparser input embeddings lead to greater model robustness, while denser inputs result in decreased robustness as shown in Finding 3 in Section \ref{results}.
\vspace{-5pt}

\subsection{Generalization to comprehensive transformer architecture}
\vspace{-5pt}
We prove that our framework can be extended for all types of transformers including encoder-only (BERT, RoBERTa, ELECTRA), decoder-only (GPT2), and encoder-decoder (BART) since the RMSE of interpolation ASR prediction is good for not only encoder-only transformers (range from 0.031 to 0.070) but also decoder-only and encoder-decoder transformers (0.025 and 0.028 respectively), shown in Table \ref{extend_table}.
\vspace{-5pt}

\subsection{Future Directions}
\label{future}

Thanks to the high accuracy (about 0.025 in RMSE) of the proposed robustness predictor, it can also be considered an influence function \cite{koh2017understanding} for robustness. Like other applications of influence function \cite{chhabra2023data,guo2021fastif,ladhak-etal-2023-contrastive}, robustness predictor is promising to be used for selecting or pruning data, robustness attribution, and data debugging to make the model more robust.
\vspace{-5pt}
\section{Conclusion}
\vspace{-5pt}
In this paper, we introduce an approach to correlate the adversarial robustness of transformer models fine-tuned on new downstream datasets. By learning a lightweight regression-based robustness predictor on a taxonomy of 13 features of a fine-tuning dataset, we empirically demonstrate that our framework can effectively predict the model robustness in both interpolation and extrapolation settings with a significant speedup in inference. 
\newpage
\pagebreak
\section*{Limitations}
\label{limit_n_future}

Although we try our best to demonstrate that our robustness evaluation toolkit can be used in practice, there are still limitations in how we design the framework. One such is that the process of fine-tuning a target transformer model does not take too much time and can be incorporated as additional signals to our algorithm. Such signals may help improve the robustness prediction performance and still ensure fast runtime. However, this approach will introduce confounding factors, and hence cannot help fully interpret the influence of fine-tuning data on model robustness, which is the main focus of this work.

Like any other ``first work'', this research direction is in its infancy. Its novelty will come with early limitations that cannot be fully resolved in one single work, and thus call for further investigations from the community. At this stage, in practice, we recommend this as an additional fast interpretable toolkit to understand and evaluate the robustness of transformer models.

\bibliography{custom}

\begin{thebibliography}{38}
\expandafter\ifx\csname natexlab\endcsname\relax\def\natexlab#1{#1}\fi

\bibitem[{Ashley(2019)}]{ashley2019brief}
Kevin~D Ashley. 2019.
\newblock A brief history of the changing roles of case prediction in ai and law.
\newblock \emph{Law Context: A Socio-Legal Journal}, 36:93.

\bibitem[{Barbieri et~al.(2020)Barbieri, Camacho-Collados, Espinosa~Anke, and Neves}]{barbieri-etal-2020-tweeteval}
Francesco Barbieri, Jose Camacho-Collados, Luis Espinosa~Anke, and Leonardo Neves. 2020.
\newblock \href {https://doi.org/10.18653/v1/2020.findings-emnlp.148} {{T}weet{E}val: Unified benchmark and comparative evaluation for tweet classification}.
\newblock In \emph{Findings of the Association for Computational Linguistics: EMNLP 2020}, pages 1644--1650, Online. Association for Computational Linguistics.

\bibitem[{Casanueva et~al.(2020)Casanueva, Tem{\v{c}}inas, Gerz, Henderson, and Vuli{\'c}}]{casanueva-etal-2020-efficient}
I{\~n}igo Casanueva, Tadas Tem{\v{c}}inas, Daniela Gerz, Matthew Henderson, and Ivan Vuli{\'c}. 2020.
\newblock \href {https://doi.org/10.18653/v1/2020.nlp4convai-1.5} {Efficient intent detection with dual sentence encoders}.
\newblock In \emph{Proceedings of the 2nd Workshop on Natural Language Processing for Conversational AI}, pages 38--45, Online. Association for Computational Linguistics.

\bibitem[{Cer et~al.(2018)Cer, Yang, Kong, Hua, Limtiaco, John, Constant, Guajardo-Cespedes, Yuan, Tar et~al.}]{cer2018universal}
Daniel Cer, Yinfei Yang, Sheng-yi Kong, Nan Hua, Nicole Limtiaco, Rhomni~St John, Noah Constant, Mario Guajardo-Cespedes, Steve Yuan, Chris Tar, et~al. 2018.
\newblock Universal sentence encoder.
\newblock \emph{arXiv preprint arXiv:1803.11175}.

\bibitem[{Chhabra et~al.(2023)Chhabra, Li, Mohapatra, and Liu}]{chhabra2023data}
Anshuman Chhabra, Peizhao Li, Prasant Mohapatra, and Hongfu Liu. 2023.
\newblock "what data benefits my classifier?" enhancing model performance and interpretability through influence-based data selection.
\newblock In \emph{International Conference on Learning Representations}.

\bibitem[{Clark et~al.(2020)Clark, Luong, Le, and Manning}]{clark2020electra}
Kevin Clark, Minh-Thang Luong, Quoc~V Le, and Christopher~D Manning. 2020.
\newblock Electra: Pre-training text encoders as discriminators rather than generators.
\newblock In \emph{International Conference on Learning Representations}.

\bibitem[{Devlin et~al.(2019)Devlin, Chang, Lee, and Toutanova}]{devlin-etal-2019-bert}
Jacob Devlin, Ming-Wei Chang, Kenton Lee, and Kristina Toutanova. 2019.
\newblock \href {https://doi.org/10.18653/v1/N19-1423} {{BERT}: Pre-training of deep bidirectional transformers for language understanding}.
\newblock In \emph{Proceedings of the 2019 Conference of the North {A}merican Chapter of the Association for Computational Linguistics: Human Language Technologies, Volume 1 (Long and Short Papers)}, pages 4171--4186, Minneapolis, Minnesota. Association for Computational Linguistics.

\bibitem[{Gao et~al.(2018)Gao, Lanchantin, Soffa, and Qi}]{gao2018black}
Ji~Gao, Jack Lanchantin, Mary~Lou Soffa, and Yanjun Qi. 2018.
\newblock Black-box generation of adversarial text sequences to evade deep learning classifiers.
\newblock In \emph{2018 IEEE Security and Privacy Workshops (SPW)}, pages 50--56. IEEE.

\bibitem[{Goodfellow et~al.(2015)Goodfellow, Shlens, and Szegedy}]{goodfellow2015explaining}
Ian~J Goodfellow, Jonathon Shlens, and Christian Szegedy. 2015.
\newblock Explaining and harnessing adversarial examples.
\newblock In \emph{International Conference on Learning Representations}.

\bibitem[{Grinsztajn et~al.(2022)Grinsztajn, Oyallon, and Varoquaux}]{grinsztajn2022tree}
L{\'e}o Grinsztajn, Edouard Oyallon, and Ga{\"e}l Varoquaux. 2022.
\newblock Why do tree-based models still outperform deep learning on typical tabular data?
\newblock \emph{Advances in Neural Information Processing Systems}, 35:507--520.

\bibitem[{Guo et~al.(2021)Guo, Rajani, Hase, Bansal, and Xiong}]{guo2021fastif}
Han Guo, Nazneen Rajani, Peter Hase, Mohit Bansal, and Caiming Xiong. 2021.
\newblock Fastif: Scalable influence functions for efficient model interpretation and debugging.
\newblock In \emph{Proceedings of the 2021 Conference on Empirical Methods in Natural Language Processing}, pages 10333--10350.

\bibitem[{Han et~al.(2024)Han, Ren, Nguyen, Nguyen, Ghosh, and Ho}]{han2024designing}
Xing Han, Tongzheng Ren, Tan Nguyen, Khai Nguyen, Joydeep Ghosh, and Nhat Ho. 2024.
\newblock Designing robust transformers using robust kernel density estimation.
\newblock \emph{Advances in Neural Information Processing Systems}, 36.

\bibitem[{Jia and Liang(2017)}]{jia-liang-2017-adversarial}
Robin Jia and Percy Liang. 2017.
\newblock \href {https://doi.org/10.18653/v1/D17-1215} {Adversarial examples for evaluating reading comprehension systems}.
\newblock In \emph{Proceedings of the 2017 Conference on Empirical Methods in Natural Language Processing}, pages 2021--2031, Copenhagen, Denmark. Association for Computational Linguistics.

\bibitem[{Jin et~al.(2020)Jin, Jin, Zhou, and Szolovits}]{Jin_Jin_Zhou_Szolovits_2020}
Di~Jin, Zhijing Jin, Joey~Tianyi Zhou, and Peter Szolovits. 2020.
\newblock \href {https://doi.org/10.1609/aaai.v34i05.6311} {Is bert really robust? a strong baseline for natural language attack on text classification and entailment}.
\newblock \emph{Proceedings of the AAAI Conference on Artificial Intelligence}, 34(05):8018--8025.

\bibitem[{Keung et~al.(2020)Keung, Lu, Szarvas, and Smith}]{keung-etal-2020-multilingual}
Phillip Keung, Yichao Lu, Gy{\"o}rgy Szarvas, and Noah~A. Smith. 2020.
\newblock \href {https://doi.org/10.18653/v1/2020.emnlp-main.369} {The multilingual {A}mazon reviews corpus}.
\newblock In \emph{Proceedings of the 2020 Conference on Empirical Methods in Natural Language Processing (EMNLP)}, pages 4563--4568, Online. Association for Computational Linguistics.

\bibitem[{Koh and Liang(2017)}]{koh2017understanding}
Pang~Wei Koh and Percy Liang. 2017.
\newblock Understanding black-box predictions via influence functions.
\newblock In \emph{Proceedings of the 34th International Conference on Machine Learning-Volume 70}, pages 1885--1894.

\bibitem[{Ladhak et~al.(2023)Ladhak, Durmus, and Hashimoto}]{ladhak-etal-2023-contrastive}
Faisal Ladhak, Esin Durmus, and Tatsunori Hashimoto. 2023.
\newblock \href {https://doi.org/10.18653/v1/2023.acl-long.643} {Contrastive error attribution for finetuned language models}.
\newblock In \emph{Proceedings of the 61st Annual Meeting of the Association for Computational Linguistics (Volume 1: Long Papers)}, pages 11482--11498, Toronto, Canada. Association for Computational Linguistics.

\bibitem[{Lehmann et~al.(2015)Lehmann, Isele, Jakob, Jentzsch, Kontokostas, Mendes, Hellmann, Morsey, Van~Kleef, Auer et~al.}]{lehmann2015dbpedia}
Jens Lehmann, Robert Isele, Max Jakob, Anja Jentzsch, Dimitris Kontokostas, Pablo~N Mendes, Sebastian Hellmann, Mohamed Morsey, Patrick Van~Kleef, S{\"o}ren Auer, et~al. 2015.
\newblock Dbpedia--a large-scale, multilingual knowledge base extracted from wikipedia.
\newblock \emph{Semantic web}, 6(2):167--195.

\bibitem[{Lewis et~al.(2020)Lewis, Liu, Goyal, Ghazvininejad, Mohamed, Levy, Stoyanov, and Zettlemoyer}]{lewis2020bart}
Mike Lewis, Yinhan Liu, Naman Goyal, Marjan Ghazvininejad, Abdelrahman Mohamed, Omer Levy, Veselin Stoyanov, and Luke Zettlemoyer. 2020.
\newblock Bart: Denoising sequence-to-sequence pre-training for natural language generation, translation, and comprehension.
\newblock In \emph{Proceedings of the 58th Annual Meeting of the Association for Computational Linguistics}, pages 7871--7880.

\bibitem[{Li et~al.(2019)Li, Ji, Du, Li, and Wang}]{li2019textbugger}
Jinfeng Li, Shouling Ji, Tianyu Du, Bo~Li, and Ting Wang. 2019.
\newblock Textbugger: Generating adversarial text against real-world applications.
\newblock In \emph{Proceedings 2019 Network and Distributed System Security Symposium}. Internet Society.

\bibitem[{Li et~al.(2020)Li, Ma, Guo, Xue, and Qiu}]{li-etal-2020-bert-attack}
Linyang Li, Ruotian Ma, Qipeng Guo, Xiangyang Xue, and Xipeng Qiu. 2020.
\newblock \href {https://doi.org/10.18653/v1/2020.emnlp-main.500} {{BERT}-{ATTACK}: Adversarial attack against {BERT} using {BERT}}.
\newblock In \emph{Proceedings of the 2020 Conference on Empirical Methods in Natural Language Processing (EMNLP)}, pages 6193--6202, Online. Association for Computational Linguistics.

\bibitem[{Liu et~al.(2019)Liu, Ott, Goyal, Du, Joshi, Chen, Levy, Lewis, Zettlemoyer, and Stoyanov}]{liu2019RoBERTa}
Yinhan Liu, Myle Ott, Naman Goyal, Jingfei Du, Mandar Joshi, Danqi Chen, Omer Levy, Mike Lewis, Luke Zettlemoyer, and Veselin Stoyanov. 2019.
\newblock Roberta: A robustly optimized bert pretraining approach.
\newblock \emph{arXiv preprint arXiv:1907.11692}.

\bibitem[{Malzer and Baum(2020)}]{malzer2020hybrid}
Claudia Malzer and Marcus Baum. 2020.
\newblock A hybrid approach to hierarchical density-based cluster selection.
\newblock In \emph{2020 IEEE international conference on multisensor fusion and integration for intelligent systems (MFI)}, pages 223--228. IEEE.

\bibitem[{Mao et~al.(2022)Mao, Qi, Chen, Li, Duan, Ye, He, and Xue}]{mao2022towards}
Xiaofeng Mao, Gege Qi, Yuefeng Chen, Xiaodan Li, Ranjie Duan, Shaokai Ye, Yuan He, and Hui Xue. 2022.
\newblock Towards robust vision transformer.
\newblock In \emph{Proceedings of the IEEE/CVF Conference on Computer Vision and Pattern Recognition}, pages 12042--12051.

\bibitem[{Pal et~al.(2024)Pal, Sulam, and Vidal}]{pal2024adversarial}
Ambar Pal, Jeremias Sulam, and Ren{\'e} Vidal. 2024.
\newblock Adversarial examples might be avoidable: The role of data concentration in adversarial robustness.
\newblock \emph{Advances in Neural Information Processing Systems}, 36.

\bibitem[{Radford et~al.(2019)Radford, Wu, Child, Luan, Amodei, Sutskever et~al.}]{radford2019language}
Alec Radford, Jeffrey Wu, Rewon Child, David Luan, Dario Amodei, Ilya Sutskever, et~al. 2019.
\newblock \href {http://www.persagen.com/files/misc/radford2019language.pdf} {Language models are unsupervised multitask learners}.
\newblock \emph{OpenAI blog}, page~9.

\bibitem[{Ren et~al.(2019)Ren, Deng, He, and Che}]{ren-etal-2019-generating}
Shuhuai Ren, Yihe Deng, Kun He, and Wanxiang Che. 2019.
\newblock \href {https://doi.org/10.18653/v1/P19-1103} {Generating natural language adversarial examples through probability weighted word saliency}.
\newblock In \emph{Proceedings of the 57th Annual Meeting of the Association for Computational Linguistics}, pages 1085--1097, Florence, Italy. Association for Computational Linguistics.

\bibitem[{Rodr{\'\i}guez~Cardona et~al.(2021)Rodr{\'\i}guez~Cardona, Janssen, Guhr, Breitner, and Milde}]{rodriguez2021matter}
Davinia Rodr{\'\i}guez~Cardona, Antje Janssen, Nadine Guhr, Michael~H Breitner, and Julian Milde. 2021.
\newblock A matter of trust? examination of chatbot usage in insurance business.
\newblock In \emph{Proceedings of the 54th Hawaii International Conference on System Sciences}, Maui, Hawaii.

\bibitem[{Sanz-Urquijo et~al.(2022)Sanz-Urquijo, Fosch-Villaronga, and Lopez-Belloso}]{sanz2022disconnect}
B~Sanz-Urquijo, E~Fosch-Villaronga, and M~Lopez-Belloso. 2022.
\newblock The disconnect between the goals of trustworthy ai for law enforcement and the eu research agenda.
\newblock \emph{AI and Ethics}, pages 1--12.

\bibitem[{Si et~al.(2021)Si, Zhang, Qi, Liu, Wang, Liu, and Sun}]{si-etal-2021-better}
Chenglei Si, Zhengyan Zhang, Fanchao Qi, Zhiyuan Liu, Yasheng Wang, Qun Liu, and Maosong Sun. 2021.
\newblock \href {https://doi.org/10.18653/v1/2021.findings-acl.137} {Better robustness by more coverage: Adversarial and mixup data augmentation for robust finetuning}.
\newblock In \emph{Findings of the Association for Computational Linguistics: ACL-IJCNLP 2021}, pages 1569--1576, Online. Association for Computational Linguistics.

\bibitem[{Sun et~al.(2020)Sun, Hashimoto, Yin, Asai, Li, Yu, and Xiong}]{sun2020adv}
Lichao Sun, Kazuma Hashimoto, Wenpeng Yin, Akari Asai, Jia Li, Philip Yu, and Caiming Xiong. 2020.
\newblock Adv-bert: Bert is not robust on misspellings! generating nature adversarial samples on bert.
\newblock \emph{arXiv preprint arXiv:2003.04985}.

\bibitem[{Van~der Maaten and Hinton(2008)}]{van2008visualizing}
Laurens Van~der Maaten and Geoffrey Hinton. 2008.
\newblock Visualizing data using t-sne.
\newblock \emph{Journal of Machine Learning Research}, 9(11).

\bibitem[{Xu et~al.(2019)Xu, Yap, and Prabhu}]{xu2019understanding}
Joyce Xu, Dian~Ang Yap, and Vinay~Uday Prabhu. 2019.
\newblock Understanding adversarial robustness through loss landscape geometries.
\newblock In \emph{Proc. of the International Conference on Machine Learning (ICML) Workshops}, volume~18.

\bibitem[{Yu et~al.(2018)Yu, Liu, Wang, Zhao, and Chen}]{yu2018interpreting}
Fuxun Yu, Chenchen Liu, Yanzhi Wang, Liang Zhao, and Xiang Chen. 2018.
\newblock Interpreting adversarial robustness: A view from decision surface in input space.
\newblock \emph{arXiv preprint arXiv:1810.00144}.

\bibitem[{Zhang et~al.(2022{\natexlab{a}})Zhang, Zhou, Wan, Zheng, Chang, and Hsieh}]{zhang2022improving}
Cenyuan Zhang, Xiang Zhou, Yixin Wan, Xiaoqing Zheng, Kai-Wei Chang, and Cho-Jui Hsieh. 2022{\natexlab{a}}.
\newblock Improving the adversarial robustness of nlp models by information bottleneck.
\newblock In \emph{Findings of the Association for Computational Linguistics: ACL 2022}, pages 3588--3598.

\bibitem[{Zhang et~al.(2015)Zhang, Zhao, and LeCun}]{zhang2015character}
Xiang Zhang, Junbo Zhao, and Yann LeCun. 2015.
\newblock Character-level convolutional networks for text classification.
\newblock \emph{Advances in Neural Information Processing Systems}, 28.

\bibitem[{Zhang et~al.(2022{\natexlab{b}})Zhang, Pan, Tan, and Kan}]{zhang2022interpreting}
Yunxiang Zhang, Liangming Pan, Samson Tan, and Min-Yen Kan. 2022{\natexlab{b}}.
\newblock Interpreting the robustness of neural nlp models to textual perturbations.
\newblock In \emph{Findings of the Association for Computational Linguistics: ACL 2022}, pages 3993--4007.

\bibitem[{Zheng et~al.(2023)Zheng, Xi, Liu, Lai, Gui, Zhang, Huang, Ma, Shan, and Ge}]{zheng-etal-2023-characterizing}
Rui Zheng, Zhiheng Xi, Qin Liu, Wenbin Lai, Tao Gui, Qi~Zhang, Xuanjing Huang, Jin Ma, Ying Shan, and Weifeng Ge. 2023.
\newblock \href {https://doi.org/10.18653/v1/2023.findings-acl.146} {Characterizing the impacts of instances on robustness}.
\newblock In \emph{Findings of the Association for Computational Linguistics: ACL 2023}, pages 2314--2332, Toronto, Canada. Association for Computational Linguistics.

\end{thebibliography}

\clearpage
\appendix

\section{Reproducibility}
\label{sec:appendix}
\subsection{Notations}
Following are the symbols used throughout our work.
\begin{itemize}
    \item $\mathcal{X}$: embedding space

    \item $X$: input sentence

    \item $\mathcal{Y}$: labels

    \item $N$: total number of samples in train data

    \item $\mathcal{C}_{n}$: naive classifier

    \item $T$: a data set with pairs of a text and a label, ($x,y$)

    \item $\mathcal{T}$: tokenizer

    \item $\mathcal{M}$: NLP classifier

    \item $\mathcal{A}$: attack success rate

    \item $\mathcal{F}$: features of train data

    \item $\mathcal{P}$: ASR predictor
\end{itemize}

\subsection{Feature Engineering}

\noindent\textbf{Embedding Distribution.} While the mean distance between classes, Fisher's discriminant ratio, and Calinski-Harabasz Index based on the labels of the inputs are used to measure the separation between classes, the number of clusters and the Davies-Bouldin Index is used to measure the density or sparseness of the embedding space. For mapping input text into multidimensional space, we use a pre-trained transformer-based Universal Sentence Encoder \cite{cer2018universal}.

\noindent$\blacksquare$ Regrading indicators for class separation, let denote the vectors mapped from input sentences to an embedding space by the Universal Sentence Encoder \cite{cer2018universal} $\mathcal{X}=\{x_{i}\}_{i=1}^{N}$, $x_{i} \in \mathbb{R}^{1\times512}$ and $\mathcal{Y}=\{y_{i}\}_{i=1}^{N}$ are their labels. Vectors with the same label will be classified into the same clusters. $\mathcal{C}=\{C_{i},N_{i},m_{i}\}_{i=1}^{K}$ is an array of clusters in the embedding space where $C_{i}$, $N_{i}$, and $m_{i}$ are a set of indexes in the $i^{th}$ cluster, the number of vectors of the $i^{th}$ cluster, and the center of the $i^{th}$ cluster respectively. $K$ is the number of clusters or possible labels in the training dataset. Since $m_{i}$ is the center of the $i^{th}$ cluster, the following formula holds:
\begin{equation*}
    m_{i}=\frac{1}{N_{i}} \displaystyle\sum_{j\in C_{i}} x_{j}
\end{equation*}
Similarly, $\{C,N,m\}$ represents the cluster covering all vectors, the number of vectors, and the global centroid. Hence, we have,
\begin{equation*}
    m=\frac{1}{N} \displaystyle\sum_{j\in C} x_{j}
\end{equation*}
Let denote $r_{i}$, $r$, $d_{ij}$ the average distance between each point of the $i^{th}$ cluster and the centroid of that cluster, also known as cluster diameter or intra-cluster distance, the global diameter and the distance between $i^{th}$ and $j^{th}$ cluster centroids, also known as inter-cluster distance. 
    \[
    \begin{array}{l}
    {\qquad\qquad r_{i}=\frac{1}{N_{i}}\displaystyle\sum_{i\in C_{i}} (x_{i}-m_{i})(x_{i}-m_{i})^{T}}\\
    {\qquad\qquad r=\frac{1}{N}\displaystyle\sum_{i\in C} (x_{i}-m)(x_{i}-m)^{T}} \\
    {\qquad\qquad d_{ij}=(m_{i}-m_{j})(m_{i}-m_{j})^{T}}
    \end{array}
    \]
The formulas for the Mean Distance between classes, Fisher's Discriminant Ratio, and Calinski-Harabasz Index are expressed as follows:
\begin{itemize}
    \item \textit{Mean Distance between classes (MD)}: This indicator calculates the average distance between the means of different classes in the input space. A larger value indicates that the means of different classes are further apart, which implies a higher degree of separation between classes.
    \begin{equation*}
        MD=2\times\frac{\displaystyle\sum_{i,j}d_{ij}}{N(N-1)}
    \end{equation*}
    \item \textit{Fisher's Discriminant Ratio (FDR)}: This metric measures the ratio of the variance between classes to the variance within classes. A larger value indicates a higher degree of separation between classes.
    \begin{equation*}
        FR=\frac{S_{B}}{S_{W}},
    \end{equation*}
    where
    \begin{equation*}
        \begin{array}{l}{{\qquad\qquad\displaystyle S_{W}=\sum_{i=1}^{K} S_{i}\qquad}}\\ 
        {\qquad\qquad S_{i}=\displaystyle\sum N_{i}\times r_{i}}\\
        {S_{B}=\displaystyle\sum_{k=1}^{K}N_{k}(m_{k}-m)(m_{k}-m)^{T}}
        \end{array}
    \end{equation*}
    \item \textit{Calinski-Harabasz Index (CHI)}: The Calinski-Harabasz index also known as the Variance Ratio Criterion, is the ratio of the sum of between-clusters dispersion and of inter-cluster dispersion for all clusters, the higher the score, the better the performances.
    \begin{equation*}
        CHI = \frac{S_{W}}{S_{B}}\times \frac{N-K}{K-1}
    \end{equation*}
\end{itemize}
We next describe in detail the formula for the features listed in Fig. \ref{taxanomy}. Features are divided into 4 main groups, namely Embedding, Distribution of Labels, Learning Ability of a Surrogate Model, and Dataset Statistics.

\noindent$\blacksquare$ For the indicators for clustering, the notations are the same as the case for class separation except that the vectors are clustered based on the HDBSCAN \cite{malzer2020hybrid} algorithm instead of being based on their labels. Hence, $K$ now is the number of clusters obtained from the HDBSCAN \cite{malzer2020hybrid} algorithm.

\begin{itemize}[leftmargin=\dimexpr\parindent-0.5\labelwidth\relax,noitemsep]
    \item \textit{Number of clusters}: This indicates how vectors in the high-dimensional space are distributed. There are $K$ clusters of vectors.
    \item \textit{Davies-Bouldin Index (DBI)}: The score is defined as the average similarity measure of each cluster with its most similar cluster, where similarity is the ratio of within-cluster distances to between-cluster distances. Thus, clusters that are farther apart and less dispersed will result in a better score. The minimum score is zero, with lower values indicating better clustering. The Davis-Bouldin Index is defined as:
    \begin{equation*}
        DBI={\frac{1}{k}}\sum_{i=1}^{k}\operatorname*{max}_{i\neq j}R_{i j},
    \end{equation*}
    where
    \begin{equation*}
        R_{i j}=\frac{r_{i}+r_{j}}{d_{ij}}
    \end{equation*}
\end{itemize}

\noindent\textbf{Label Distribution.} Let denote $\mathcal{Y}$ as a random variable representing possible labels in the training dataset $\mathcal{S}_{train}$.
\begin{itemize}[leftmargin=\dimexpr\parindent-0.5\labelwidth\relax,noitemsep]
    \item \textit{Pearson Median Skewness (PMS)}: The sign indicates the direction of the skewness. The coefficient compares the distribution of the sample to that of a normal distribution. The greater the value, the greater the deviation from the normal distribution. A value of 0 denotes that there is no skewness at all. A big negative value indicates that the distribution is skewed. A high positive value indicates that the distribution is biased to the right.
    \begin{equation*}
        PMS_={\frac{3(\bar{\mathcal{Y}}-M d)}{s}},
    \end{equation*}
    where $\bar{\mathcal{Y}}$, $M d$, and $s$ are respectively mean, median, and variance of the distribution of labels.
    \item \textit{Kurtosis (Kurt)}: Kurtosis is a measure of the peakedness or flatness of a distribution. A distribution with kurtosis equal to 3 is considered to be \textit{mesokurtic} (i.e., having a normal distribution), while a distribution with kurtosis greater than 3 is considered to be \textit{leptokurtic} (i.e., having a sharper peak and fatter tails) and a distribution with kurtosis less than 3 is considered to be \textit{platykurtic} (i.e., having a flatter peak and thinner tails).
    \begin{equation*}
        Kurt=\frac{\operatorname{E}\left[(\mathcal{Y}-\bar{\mathcal{Y}})^{4}\right]}{(\operatorname{E}[(\mathcal{Y}-\bar{\mathcal{Y}})^{2}])^{2}}
    \end{equation*}
\end{itemize}

\noindent\textbf{Surrogate Model's Learnability.}
\begin{itemize}[leftmargin=\dimexpr\parindent-0.5\labelwidth\relax,noitemsep]
    \item \textit{Misclassification Rate (MCR)}: We create a naive classifier $\mathcal{C}_{n}:X\rightarrow \mathcal{Y}$ that turns text, $X$, into predicted classes, $\mathcal{Y}$, and analyze the performance of this classifier because we think misclassification is related to its robustness. \cite{sun2020adv} shows that typos affect the robustness of the transformer-based model, so we choose the character-based CNN \cite{zhang2015character} model for $\mathcal{C}_{n}$ because it can exploit character-level properties. Suppose classfier $\mathcal{C}_{n}$ turns a set $T$ of text $x_{i}$ with true class $y_{i}$ into a predicted class $\hat y_{i}=\mathcal{C}_{n}(x_{i})$. Misclassification Rate is expressed by the following formula:
    \begin{equation*}
        MCR=\frac{|\{\hat{y_{i}}\neq y_{i}|(x_{i},y_{i}) \in T\}|}{|T|}
    \end{equation*}
\end{itemize}

\noindent\textbf{Token-Based Statistics.} We use a tokenizer $\mathcal{T}$, namely Bert-base-cased tokenizer \cite{devlin-etal-2019-bert}, to convert a sentence into an array of tokens. The notation $X$ is a text set in the training dataset $\mathcal{S}_{train}$. Tokenizer $\mathcal{T}$ converts each text $x_i \in X$ into a list of $M_{i}$ tokens $\{t_{ij}\}_{j=1}^{M_{i }}$, $\mathcal{T}:t_i \rightarrow \{t_{ij}\}_{j=1}^{N_{i}}$. Denote $Y$ the collection of lists of tokens, so $T$ turns $X$ into $Y$. The formulas for those syntactic features are illustrated as follows:

\begin{equation}
\begin{aligned}
    \qquad\qquad \textrm{Avg. \# tokens} &= \frac{1}{|X|}\displaystyle\sum_{i=1}^{|X|} M_{i} \\
    \textrm{\# unique tokens} &= |\{t|t\in 
    \displaystyle\cup_{i=1}^{|X|} \{t_{ij}\}_{j=1}^{M_{i}} \\ 
    & \& ~t~ \textrm{exists once}\}|     \\
    \qquad\qquad \textrm{Min \# tokens} &= min(\{M_{i}\}_{i=1}^{|X|}) \\
    \qquad\qquad \textrm{Max \# tokens} &= max(\{M_{i}\}_{i=1}^{|X|})
\end{aligned}        
\end{equation}

In addition, the \textit{total number of classes} is number of possible classes of dataset $\mathcal{D}$ from which the sub-dataset $S_{train}$ is sampled

\subsection{Evaluation Metrics}

Denote $\mathcal{\hat{A}}$ and $\mathcal{A}$ the predicted ASR made by $\mathcal{P}$ and the actual ASR. The metrics we use to evaluate ASR predictors include the following:
\begin{itemize}[leftmargin=\dimexpr\parindent-0.5\labelwidth\relax,noitemsep]
    \item \textit{Root Mean Squared Error (RMSE)}: This is the square root of the mean square error (MSE), the average of the squared differences between the predicted values and the actual values.
    \begin{equation}
        RMSE = \sqrt{\frac{\sum_{i=1}^{N}\left(\mathcal{A}_{i}-\mathcal{\hat{A}}_{i}\right)^{2}}{N}},
    \end{equation} where $N$ is the number of predicted values.
    \item \textit{Mean Absolute Error (MAE)}: This is the average of the absolute differences between the predicted values and the actual values. It is less sensitive to outliers than MSE, but may not penalize large errors as heavily.
    \begin{equation}
        MAE = \frac{1}{N}\sum_{i=1}^{N} \left|\mathcal{A}_{i}-\mathcal{\hat{A}}_{i}\right|
    \end{equation}
    \item \textit{R-squared (R2)}: This is measured by the ratio between the mean squared error of a regression model and the variance of the target variable. It ranges from 0 to 1, with higher values indicating better performance.
    \begin{equation}
        R^{2}=1-\frac{\sum_{i=1}^{N}(\mathcal{A}_{i}-\mathcal{\hat{A}}_{i})^{2}}{\sum_{i=1}^{N}(\mathcal{A}_{i}-\mathcal{\bar{A}})^{2}},
    \end{equation} where $\mathcal{\bar{A}}=\frac{\sum_{i=1}^{N}\mathcal{A}_{i}}{N}$
    \item \textit{Mean Absolute Percentage Error (MAPE)}: This is the average of the absolute percentage differences between the predicted values and the actual values. It is commonly used in forecasting applications.
    \begin{equation}
        MAPE = \frac{1}{N}\sum_{i=1}^{N} \left|\frac{\mathcal{A}_{i}-\mathcal{\hat{A}}_{i}}{\mathcal{A}_{i}} \right|
    \end{equation}
    \item \textit{Explained Variance Score (EVS)}: This is the proportion of variance in the target variable that is explained by the model relative to the total variance. It ranges from 0 to 1, with higher values indicating better performance.
    \begin{equation}
        EVS = 1- \frac{Var(\mathcal{A}-\mathcal{\hat{A}})}{Var(\mathcal{A})}
    \end{equation}
\end{itemize}

\subsection{Experiment Setup}
\label{experiment_setup}
\noindent\textbf{Hardware Specifications.} 
We use one GPU NVIDIA RTX A6000 and 32 CPUs AMD Ryzen Threadripper PRO 5975WX 32-Cores for our experiment.

\subsection{Error Analysis}
We examine how features impact errors in ASR prediction. Initially, we employ the Random Forest model to predict ASR for test samples. Subsequently, test samples exhibiting errors surpassing the 70th percentile are categorized as False; otherwise, they are labeled as True. Then, logistic regression is applied to distinguish outlier test samples. Ultimately, the absolute magnitude of parameter weights from the logistic regression model is utilized to gauge feature significance. Essentially, these parameter weights indicate the degree of error influencing ASR prediction.

\noindent\textbf{Sampling Datapoints.} 

The dataset list, $\mathcal{D}_{l}$, that we have used includes AG News, Amazon Review Full, Amazon Review Polarity, DPedia, Yahoo Answers, Yelp Review Full, Amazon Review Polarity, Banking 77, Emoji-TweetEval. In $\mathcal{D}_{l}$, we divided the datasets into two groups: $\mathcal{D}_{a} = \{$ AG News, Amazon Review Full, Amazon Review Polarity, DPedia, Yahoo Answers, Yelp Review Full, Amazon Review Polarity $\}$ and $\mathcal{D}_{b}=\{$ Banking 77, Emoji-TweetEval $\}$. The datasets in $\mathcal{D}_{a}$ have label counts of 2, 4, 5, 10, and 14 and are confined to a few settings, such as Yelp reviews, news articles, Yahoo inquiries, etc while those in $\mathcal{D}_{b}$ are 77 and 22. Due to the lack of diversity within each group of label counts, datasets in $\mathcal{D}_{a}$ are kept, whereas datasets in $\mathcal{D}_{b}$ are slightly adjusted to boost contextual variety for each group of label count. When a sub-dataset $(S^i_\mathrm{train}, S^i_\mathrm{val}, S^{d}_\mathrm{test})$ in $\mathcal{\pmb{Q}}$ introduced in Algorithm \ref{alg:algorithm} is sampled, if it is in the $\mathcal{D}_{b}$, the number of classes of that sub-dataset will also be randomly converted to 2, 4, 5, 10, or 14. For example, to convert a 22-label Emoji-TweetEval dataset into a 4-label dataset, the labels from 1 to 5, from 5 to 10, from 11 to 15, and from 16 to 20 will be converted into the new labels 1, 2, 3, 4, while samples with residual labels 21, 22 will be discarded. 

After that, we sampled 500 data points of train data features and attack success rates in 72 hours. 

\noindent\textbf{BERT and ROBERTA hyperparameters.} The tokenizer has a maximum length of 512 words. The learning rate, weight decay, and warmup step are 5e-4, 0.01, and 500, respectively. We train 5 epochs in that, from our observations, it is enough for the model to converge and get good inference results on the test dataset.

\noindent\textbf{Character-level CNN.} A tokenizer converts each character in a sentence into a one-hot vector in this model. It can only be 1024 characters long. There are six CNN layers, the kernel size of the first one and the five following are 3 and 7 respectively. The first and final CNN layer is followed by a 3 kernel size 1D pooling layer.

\noindent\textbf{HDBSCAN.} The minimum cluster size is five. The metric is Euclidean.

\noindent\textbf{Gradient Boosting Regressor.} The learning rate, maximum bin, and number of estimators are 0.05, 400, and 5000, respectively.

\begin{table*}[tb]
\centering
\footnotesize
\begin{tabular}{llrrrrrr}
\toprule
{} & \multicolumn{1}{c}{\multirow{2}{*}{\textbf{Metric}}} & \multicolumn{3}{c}{\textbf{Interpolation}} & \multicolumn{3}{c}{\textbf{Extrapolation}} \\
\cmidrule(lr){3-5} \cmidrule(lr){6-8}
{} & {} & \multicolumn{1}{c}{\textbf{GB}} & \multicolumn{1}{c}{\textbf{LR}} & \multicolumn{1}{c}{\textbf{RF}} & \multicolumn{1}{c}{\textbf{GB}} & \multicolumn{1}{c}{\textbf{LR}} & \multicolumn{1}{c}{\textbf{RF}} \\
\cmidrule(lr){1-8}
\multirow{5}{*}{\rotatebox[origin=c]{90}{\textbf{BERT}}}        & RMSE${\downarrow}$ & $0.059\pm0.000$ & $0.072\pm0.000$ & $\bf0.055\pm0.000$ & $0.169\pm0.005$ & $0.063\pm0.003$ & $\bf0.063\pm0.001$ \\
\textbf{}        & $R^2$${\uparrow}$   & $0.892\pm0.006$ & $0.841\pm0.007$ & $\bf0.904\pm0.005$ & $0.394\pm0.177$ & $0.871\pm0.122$ & $\bf0.885\pm0.033$ \\
{}    & MAE${\downarrow}$  & $0.040\pm0.000$ & $0.053\pm0.000$ & $\bf0.037\pm0.000$ & $0.128\pm0.006$ & $\bf0.040\pm0.001$ & $0.045\pm0.000$ \\
\textbf{}        & EVS${\uparrow}$  & $0.895\pm0.006$ & $0.846\pm0.007$ & $\bf0.907\pm0.005$ & $0.522\pm0.089$ & $0.892\pm0.060$ & $\bf0.908\pm0.021$ \\
\textbf{}        & MAPE${\downarrow}$ & $0.077\pm0.001$ & $0.101\pm0.001$ & $\bf0.071\pm0.000$ & $0.278\pm0.020$ & $\bf0.086\pm0.006$ & $0.102\pm0.004$ \\ 
\cmidrule(lr){1-8}
\multirow{5}{*}{\rotatebox[origin=c]{90}{\textbf{RoBERTa}}}        & RMSE${\downarrow}$ & $0.037\pm0.000$ & $0.056\pm0.000$ & $\bf0.031\pm0.000$ & $0.206\pm0.005$ & $0.073\pm0.003$ & $\bf0.061\pm0.001$ \\
\textbf{}        & $R^2$${\uparrow}$   & $0.959\pm0.000$ & $0.907\pm0.001$ & $\bf0.972\pm0.000$ & $0.139\pm0.145$ & $0.829\pm0.205$ & $\bf0.900\pm0.019$ \\
\textbf{}    & MAE${\downarrow}$  & $0.028\pm0.000$ & $0.044\pm0.000$ & $\bf0.025\pm0.000$ & $0.176\pm0.006$ & $\bf0.042\pm0.001$ & $0.044\pm0.000$ \\
\textbf{}        & EVS${\uparrow}$ & $0.961\pm0.000$ & $0.911\pm0.001$ & $\bf0.972\pm0.000$ & $0.309\pm0.109$ & $0.846\pm0.153$ & $\bf0.922\pm0.010$ \\
\textbf{}        & MAPE${\downarrow}$ & $0.054\pm0.000$ & $0.083\pm0.000$ & $\bf0.048\pm0.000$ & $0.385\pm0.032$ & $\bf0.083\pm0.003$ & $0.095\pm0.004$ \\ 
\cmidrule(lr){1-8}
\multirow{5}{*}{\rotatebox[origin=c]{90}{\textbf{ELECTRA}}}        & RMSE${\downarrow}$ & $0.107\pm0.001$ & $0.084\pm0.002$ & $\bf0.070\pm0.001$ & $0.135\pm0.004$ & $0.148\pm0.009$ & $\bf0.073\pm0.000$ \\
\textbf{}        & $R^2$${\uparrow}$   & $0.411\pm0.492$ & $0.635\pm0.194$ & $\bf0.686\pm0.490$ & $0.450\pm0.240$ & $0.348\pm0.694$ & $\bf0.864\pm0.007$ \\
{}    & MAE${\downarrow}$  & $0.083\pm0.001$ & $0.057\pm0.001$ & $\bf0.047\pm0.000$ & $0.100\pm0.005$ & $0.064\pm0.000$ & $\bf0.039\pm0.000$ \\
\textbf{}        & EVS${\uparrow}$  & $0.505\pm0.293$ & $0.677\pm0.152$ & $\bf0.729\pm0.326$ & $0.513\pm0.174$ & $0.361\pm0.671$ & $\bf0.870\pm0.005$ \\
\textbf{}        & MAPE${\downarrow}$ & $0.151\pm0.006$ & $0.105\pm0.004$ & $\bf0.084\pm0.003$ & $0.180\pm0.012$ & $0.129\pm0.002$ & $\bf0.077\pm0.000$ \\ 
\cmidrule(lr){1-8}
\multirow{5}{*}{\rotatebox[origin=c]{90}{\textbf{GPT2}}}        & RMSE${\downarrow}$ & $0.093\pm0.002$ & $0.026\pm0.000$ & $\bf0.025\pm0.000$ & $0.110\pm0.002$ & $0.147\pm0.009$ & $\bf0.078\pm0.000$ \\
\textbf{}        & $R^2$${\uparrow}$   & $-0.468\pm37.303$ & $0.888\pm0.105$ & $\bf0.890\pm0.106$ & $0.523\pm0.135$ & $-0.013\pm3.437$ & $\bf0.794\pm0.005$ \\
\textbf{}    & MAE${\downarrow}$  & $0.067\pm0.001$ & $\bf0.020\pm0.000$ & $0.022\pm0.000$ & $0.079\pm0.002$ & $0.069\pm0.000$ & $\bf0.051\pm0.000$ \\
\textbf{}        & EVS${\uparrow}$ & $0.019\pm10.871$ & $0.911\pm0.056$ & $\bf0.913\pm0.049$ & $0.545\pm0.122$ & $0.005\pm3.314$ & $\bf0.801\pm0.005$ \\
\textbf{}        & MAPE${\downarrow}$ & $0.107\pm0.004$ & $\bf0.028\pm0.000$ & $0.030\pm0.000$ & $0.136\pm0.005$ & $0.126\pm0.001$ & $\bf0.009\pm0.000$ \\
\cmidrule(lr){1-8}
\multirow{5}{*}{\rotatebox[origin=c]{90}{\textbf{BART}}}        & RMSE${\downarrow}$ & $0.052\pm0.000$ & $0.041\pm0.001$ & $\bf0.028\pm0.000$ & $0.107\pm0.003$ & $0.070\pm0.003$ & $\bf0.068\pm0.001$ \\
\textbf{}        & $R^2$${\uparrow}$   & $0.856\pm0.014$ & $0.885\pm0.028$ & $\bf0.995\pm0.001$ & $0.423\pm0.264$ & $0.743\pm0.222$ & $\bf0.813\pm0.019$ \\
\textbf{}    & MAE${\downarrow}$  & $0.039\pm0.000$ & $0.028\pm0.000$ & $\bf0.022\pm0.000$ & $0.074\pm0.003$ & $\bf0.032\pm0.000$ & $0.036\pm0.000$ \\
\textbf{}        & EVS${\uparrow}$ & $0.875\pm0.009$ & $0.896\pm0.044$ & $\bf0.960\pm0.001$ & $0.501\pm0.124$ & $0.747\pm0.213$ & $\bf0.822\pm0.017$ \\
\textbf{}        & MAPE${\downarrow}$ & $0.070\pm0.002$ & $0.053\pm0.002$ & $\bf0.036\pm0.000$ & $0.124\pm0.005$ & $\bf0.063\pm0.001$ & $0.076\pm0.001$ \\
\bottomrule
\end{tabular}
\caption{ASR results (mean$\pm$std) under interpolation and extrapolation prediction on BERT, RoBERTa, ELECTRA, BART and GPT2 using three classifiers, namely Gradient Boosting (GB), Linear Regression (LR), and Random Forest (RF).}
\label{extend_table}
\end{table*}
\end{document}